\documentclass[nohyperref]{article}

\usepackage{microtype}
\usepackage{graphicx}
\usepackage{subfigure}
\usepackage{booktabs} % for professional tables

\usepackage{hyperref}

\usepackage[accepted]{icml2022}

\usepackage{amsmath}
\usepackage{amssymb}
\usepackage{mathtools}
\usepackage{amsthm}

\usepackage{multirow}
\usepackage{makecell}
\usepackage{amsfonts} 
\usepackage{colortbl}

\usepackage[capitalize,noabbrev]{cleveref}

\theoremstyle{plain}

\theoremstyle{definition}

\theoremstyle{remark}

\usepackage[textsize=tiny]{todonotes}

\icmltitlerunning{Transformer in Transformer as Backbone for Deep Reinforcement Learning}

\begin{document}

\twocolumn[
\icmltitle{Transformer in Transformer as Backbone for Deep Reinforcement Learning}



\icmlsetsymbol{equal}{*}

\begin{icmlauthorlist}
\icmlauthor{Hangyu Mao}{noah}
\icmlauthor{Rui Zhao}{MML}
\icmlauthor{Hao Chen}{ucas}
\icmlauthor{Jianye Hao}{noah,tianjin}
\icmlauthor{Yiqun Chen}{ucas}
\icmlauthor{Dong Li}{noah}
\icmlauthor{Junge Zhang}{sai-ucas,crise-cas}
\icmlauthor{Zhen Xiao}{pku}
\end{icmlauthorlist}

\icmlaffiliation{noah}{Noah's Ark Lab, Huawei}
\icmlaffiliation{MML}{Multimedia Laboratory, The Chinese University of Hong Kong}
\icmlaffiliation{ucas}{University of Chinese Academy of Sciences}
\icmlaffiliation{tianjin}{Tianjin University}
\icmlaffiliation{sai-ucas}{School of Artificial Intelligence, University of Chinese Academy of Sciences}
\icmlaffiliation{crise-cas}{CRISE, Institute of Automation, Chinese Academy of Sciences}
\icmlaffiliation{pku}{Peking University}

\icmlcorrespondingauthor{Hangyu Mao}{hy.mao@pku.edu.cn}

\icmlkeywords{Reinforcement Learning, Transformer, Transformer in Transformer, Machine Learning, ICML}

\vskip 0.3in
]



\printAffiliationsAndNotice{}  % leave blank if no need to mention equal contribution

\begin{abstract}
Designing better deep networks and better reinforcement learning (RL) algorithms are both important for deep RL. This work focuses on the former. Previous methods build the network with several modules like CNN, LSTM and Attention. Recent methods combine the Transformer with these modules for better performance. However, it requires tedious optimization skills to train a network composed of mixed modules, making these methods inconvenient to use in practice. In this paper, we propose to design \emph{pure Transformer-based networks} for deep RL, aiming at providing off-the-shelf backbones for both the online and offline settings. Specifically, the Transformer in Transformer (TIT) backbone is proposed, which cascades two Transformers in a very natural way: the inner one is used to process a single observation, while the outer one is responsible for processing the observation history. Combining both is expected to extract spatial-temporal representations for good decision-making. Experiments show that TIT can achieve satisfactory performance in different settings consistently.
\end{abstract}

\section{Introduction}
Deep reinforcement learning (RL) has made great progress in various domains, such as mastering video games \cite{mnih2015human}, cooling data centers \cite{evans2016deepmind}, and manipulating robotic arms \cite{kalashnikov2018scalable}. In general, a \emph{deep RL method} mainly consists of two parts: the \emph{deep network} that generates the action (and value), and the \emph{RL algorithm} that trains the deep network.

Most deep RL methods focus on the innovation of the RL algorithms. The representative methods include but are not limited to the fundamental DQN \cite{mnih2015human}, the asynchronous methods like A3C \cite{mnih2016asynchronous} and GA3C \cite{babaeizadeh2016ga3c}, the deterministic policy gradient methods like DDPG \cite{lillicrap2015continuous} and TD3 \cite{fujimoto2018addressing}, the trust region based policy optimization methods like TRPO \cite{schulman2015trust} and PPO \cite{John2017PPO}, the distributional methods like C51 \cite{bellemare2017distributional} and QR-DQN \cite{dabney2018distributional}, the model-based methods like Dreamer \cite{hafner2019dream} and Dreaming \cite{okada2021dreaming}, and even the offline methods like BCQ \cite{fujimoto2019off} and CQL \cite{kumar2020conservative}. It is fair to say that a large variety of RL algorithms can be used in practice.

In contrast, although the design of better deep networks for deep RL is a very important and long-pursuing target for the community (see Section \ref{sec:RelatedWork}), we have much fewer \textit{network modules} to choose to build the deep network. Specifically, the Convolutional Neural Network (CNN) and Multilayer Perceptron (MLP) are directly used in most cases, e.g, in DQN. When the problem is partially-observable, the Long Short-Term Memory (LSTM) \cite{greff2016lstm} and Attention are typically used, e.g., in DRQN \cite{hausknecht2015deep} and DARQN \cite{sorokin2015deep}. Recently, the Transformer \cite{vaswani2017attention} has been applied to build better networks for deep RL, resulting in some successful methods like GTrXL \cite{parisotto2020stabilizing}, Catformer \cite{davis2021catformer} and CoBERL \cite{banino2021coberl}. 

Although Transformer-based methods can achieve better performance than previous methods, it is inconvenient to use them in practice. Firstly, these methods need combine Transformers with other network modules, but figuring out the suitable way to combine them is sometimes difficult. More seriously, different modules need different optimization skills, but it is hard to master all skills for all modules, letting alone optimizing them jointly. For example, GTrXL and Catformer combine ResNet \cite{he2016deep}, Transformer-XL \cite{dai2019transformer}, Gating and MLP, while CoBERL combines ResNet, BERT \cite{kenton2019bert}, LSTM, Gating, and MLP. During model training, the authors apply a lot of optimization skills, e.g., pretraining of ResNet, special initialization for network parameters, contrastive representation learning, regularization for self-attention consistency, distributed training, and so on. It makes these methods very unstable to be optimized, and the performance varies greatly as shown by our experiments. %  entropy management, prioritized experience replay, 

Because mastering the skills to train networks composed of hybrid modules is especially hard, we ask the following question in this paper: can we design deep networks for deep RL \emph{purely} based on Transformers, so that we can \emph{use it as backbone off-the-shelf} to achieve good performance, \emph{without being disturbed by} the combination of other network modules and the tedious optimization skills?
% can we design deep networks for deep RL based on only one type of network module

%The target network module is very likely to be Transformer. The reasons are as follows.
We target at Transformers rather than other network modules due to the following reasons. Firstly, Transformer \cite{vaswani2017attention,kenton2019bert,brown2020language} and Vision Transformer \cite{dosovitskiy2020image,liu2021swin,han2021transformer} are the most outstanding breakthroughs in natural language processing (NLP) and computer vision (CV), and they have demonstrated the practical effectiveness and scalability for processing sequential words and image patches (and more generally, sequential data). Secondly, the goal of deep RL is sequential decision-making based on sequential observations (and popularly, the observation is an image), which matches the ability of Transformer and Vision Transformer exactly. Lastly, reviewing previous deep RL methods like DQN, DRQN, DARQN and these discussed in Section \ref{sec:RelatedWork}, we find that (1) their deep networks are largely inspired by the breakthroughs in NLP and CV, and that (2) both NLP and CV breakthroughs will eventually result in better deep RL performance when properly applied, so do Transformers we believe.

% In this paper, in order to properly apply Transformers as the backbone network for deep RL, 
To this end, we propose to explore two Transformer in Transformer (TIT) backbones. Specifically, both of the two TIT backbones are made up of an Inner Transformer and an Outer Transformer: the inner one is used to process a single observation at the \emph{observation patch} level to learn a good observation representation that captures important spatial information in an observation, while the outer one is responsible for processing sequential observations (or tuples of $\langle \text{return, observation, action} \rangle$ in the learning paradigm of Decision Transformer \cite{chen2021decision}) to capture important temporal information across multiple consecutive observations, and combining both is expected to extract spatial-temporal representations for better decision-making.

The two TIT backbones differ in the way how the two Transformers are cascaded. The Vanilla\_TIT shown in Figure \ref{fig:TIT_First} applies $L$ Inner Transformer blocks before $L$ Outer Transformer blocks. There is no information interaction between the two types of blocks. In contrast, the Enhanced\_TIT shown in Figure \ref{fig:TIT_Second} builds its TIT block by the inner block and the outer block, and stacks $L$ TIT blocks to form the TIT backbone. Therefore, it can fuse the spatial-temporal information in every TIT blocks, which may capture more suitable representations compared to Vanilla\_TIT. %  Compared to previous methods like GTrXL and CoBERL, it can be seen as simply replacing their ResNet in the low layers with the Inner Transformer. 

Our contributions are summarized as follows. (1) To our best knowledge, we are the first to show that pure Transformers can serve as the backbones for both the standard online and offline RL methods (e.g., PPO and CQL), as long as we design the backbones properly. This plays a similar role as ViT \cite{dosovitskiy2020image}, which proves that pure Transformer-based networks can perform well for CV tasks. Moreover, pure Transformer-based backbones bring advantages like that it needs fewer optimization skills; it is agnostic to the RL training algorithms; it can be used off-the-shelf by combining with popular RL libraries like stable-baseline3 \cite{SB3}; and it may handle the complex observations like 3D images more effectively. (2) We propose two TIT backbones. We empirically show that the best one achieves comparable or better performance than several strong baselines not only in online and offline RL settings, but also in the supervised learning paradigm recently proposed by Decision Transformer. (3) We also analyze the backbone from a lot of aspects to provide better understanding of our methods. %(4) We will open the source code to accelerate the future research. % Notably, the operations in Transformers are easier to be parallel, which presents a future opportunity to explore the scaling laws for deep RL, just as the recent successes in NLP and CV.

\section{Related Work}\label{sec:RelatedWork}
The design of better deep networks for deep RL methods has drawn relatively less attention, compared to the design of RL algorithms. In this paper, we roughly divide the traditional methods into two categories: the RL-property-consistent networks \cite{Wang2016Dueling,tamar2016value} and the larger-deeper networks \cite{Ota2020Can,sinha2020d2rl,bjorck2021towards, ota2021training}. The Dueling Network \cite{Wang2016Dueling} and Value Iteration Network \cite{tamar2016value} are the representative studies of the former type. They design a two-branch dueling network to compute the advantages of actions and a value iteration module to mimic the procedure of value iteration, respectively. They have been selected as the best papers of ICML'16 and NIPS'16, respectively. It means that designing better deep networks is an important and recognized target of the deep RL community. The latter type is largely influenced by the breakthroughs in CV community, e.g., OFENet \cite{Ota2020Can} and D2RL \cite{sinha2020d2rl} focus on designing larger and deeper networks by applying different variants of DenseNet \cite{Huang2017Densely} to stabilize training.

Recently, there is a trend of applying Transformers to deep RL, which is well surveyed by \cite{hu2022TRL}. As far as we know, the recent methods only use one Transformer to process either a single image observation independently \cite{sopov2022transformer,meng2022deep,tao2022evaluating} or observation history as a whole \cite{parisotto2020stabilizing,davis2021catformer,banino2021coberl,goulao2022pretraining}. Besides, these methods often mix several network modules, e.g, CoBERL \cite{banino2021coberl} mixes ResNet, BERT, LSTM and Gating, which makes these methods unstable to train and hard to deploy \cite{davis2021catformer}. In contrast, we apply two Transformers to process a single observation and the observation history harmoniously, and we focus on exploring whether the pure Transformer-based networks can achieve good performance for deep RL, independent of the training algorithms and the optimization skills.

There are some works applying Transformers in meta RL \cite{melo2022transformers} and model-based RL \cite{chen2021transdreamer,micheli2022transformers}. Methods like Decision Transformer \cite{chen2021decision} and Trajectory Transformer \cite{janner2021reinforcement} solve offline RL problem by training a Transformer with supervised learning. These works are orthogonal to ours, since we aim at \emph{designing} pure Transformer-based backbones for deep RL.

In CV community, some studies, e.g., TNT \cite{han2021transformer}, ViViT \cite{Arnab2021ViViT}, DualFormer \cite{Liang2022DualFormer} and COAT \cite{Yu2022CascadeTransformers}, apply two or more Transformers to handle tasks like image classification and person search. But our preliminary experiments show that these designs are too complex for RL tasks, which motivates our minimal implementation of TIT.

\section{Background}
We consider the problems that can be formulated as a Markov Decision Process (MDP), which is formally defined by a tuple $\langle S, A, T, R, \gamma \rangle$, where $S$ is the set of possible states $s \in S$; $A$ represents the set of possible actions $a \in A$; $T(s'|s,a): S \times A \times S \rightarrow [0,1]$ denotes the state transition function; $R(s,a): S \times A \rightarrow \mathbb{R}$ is the reward function; $\gamma \in [0,1]$ is the discount factor. We use $s_t$, $a_t$ and $r_t=R(s_t,a_t)$ to denote the state, action and reward at timestep $t$, respectively. Our goal is to learn a policy $\pi(a_t|s_t)$ that can maximize $\mathbb{E}[\hat{R}]$ where $\hat{R}=\Sigma^{H}_{t=0}\gamma^{t}r_t$ is the return, and $H$ is the time horizon. Reinforcement learning \cite{sutton2018reinforcement} is a popular approach to solve the MDP problems. In practice, the environment can be noisy, so we can only get an observation $o_t$, which contains partial information of the state $s_t$. We have to learn the policy based on the observation history $\langle o_{t-(K-1)}, ..., o_{t-1}, o_{t} \rangle$. This setting is called partially-observable MDP (POMDP). % \cite{spaan2012partially}. %Background about Transformers are shown in Appendix.

\begin{figure*}[!thb]
\vskip 0.2in
\begin{center}
\centerline{\includegraphics[width=1.90\columnwidth]{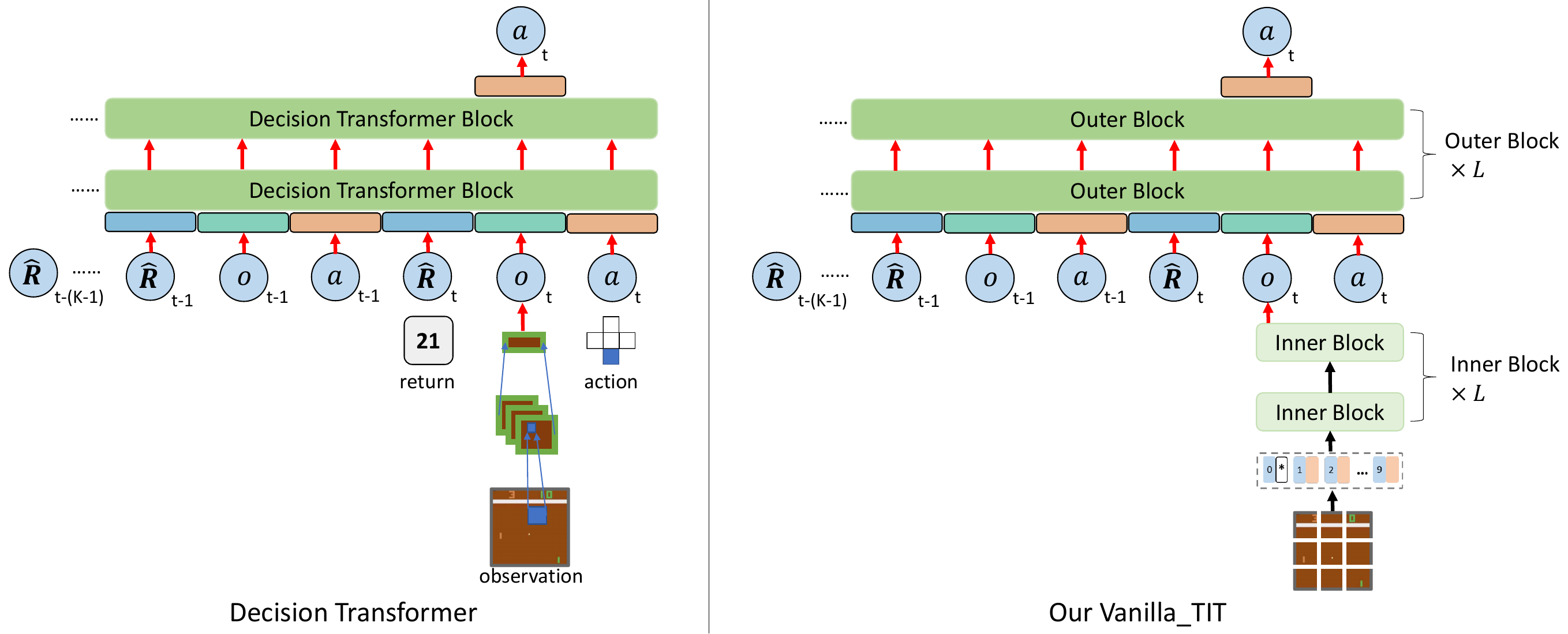}}
\caption{The architecture of the proposed Vanilla\_TIT (right). Compared with Decision Transformer (left), the Vanilla\_TIT can be regarded as simply replacing its CNN in the lower layers with the most basic Vision Transformer (i.e., the Inner Transformer). Compared with GTrXL and CoBERL, the Vanilla\_TIT can be seen as replacing their ResNet in the lower layers with the Inner Transformer, and further replacing their Transformer-XL or BERT with the Outer Transformer.}
\label{fig:TIT_First}
\end{center}
\vskip -0.2in
\end{figure*}

\begin{figure*}[!thb]
\vskip 0.2in
\begin{center}
\centerline{\includegraphics[width=1.92\columnwidth]{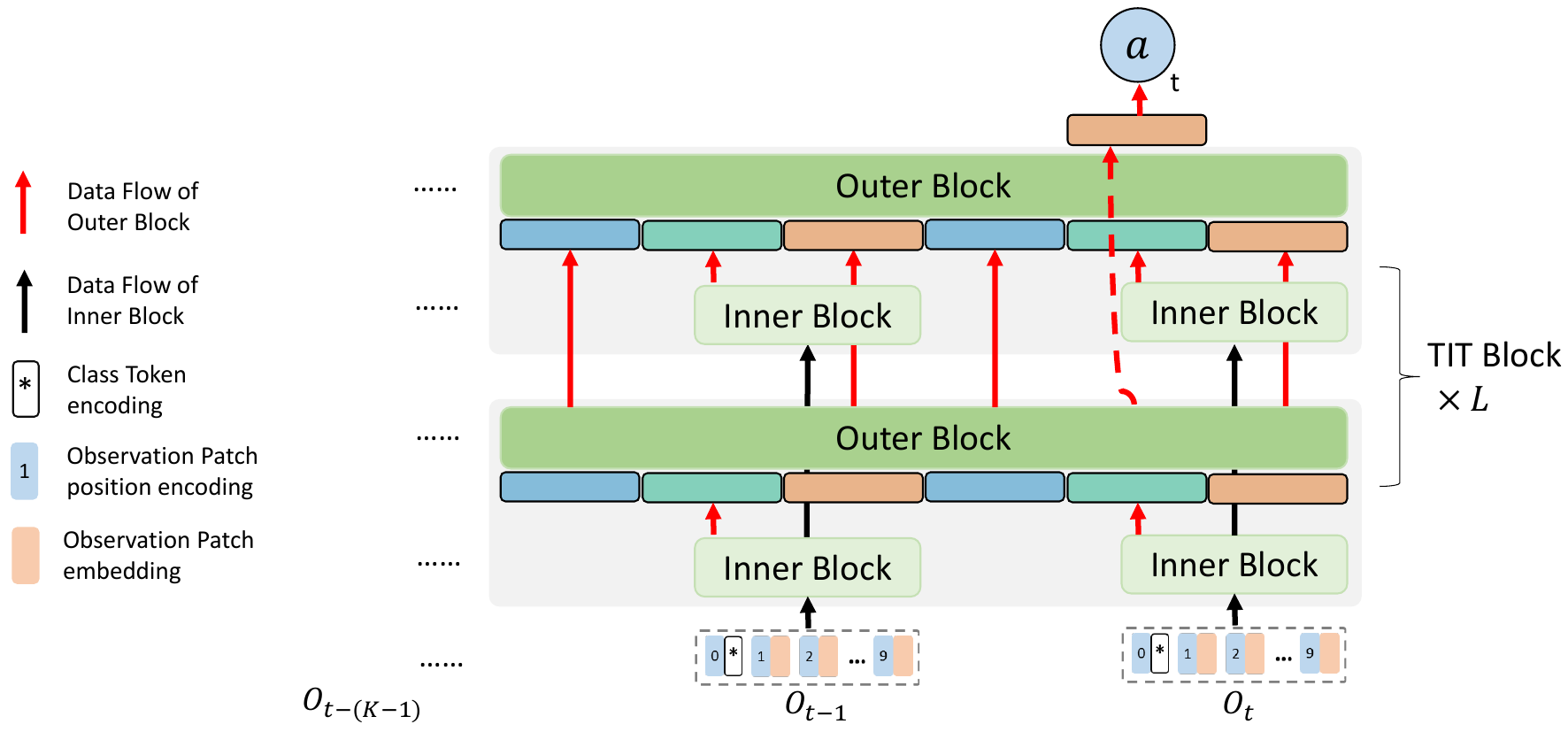}}
\caption{The architecture of the proposed Enhanced\_TIT. It not only fuses the spatial-temporal information in every TIT blocks (i.e., the gray rectangles), but also adopts a dense connection design (i.e., the red dotted lines) to direct information from each TIT blocks to the final output explicitly. Note that the inner blocks in the same layer share model parameters across different timesteps.}
\label{fig:TIT_Second}
\end{center}
\vskip -0.2in
\end{figure*}

\section{Approach}
%We propose two Transformer in Transformer (TIT) backbones for deep RL: the Vanilla\_TIT and the Enhanced\_TIT, as shown in Figure \ref{fig:TIT_First} and \ref{fig:TIT_Second}, respectively. The proposed backbones mainly consist of two Transformers: the inner one is used to process a single observation; the outer one is responsible for processing the observation history (i.e., multiple observations); combining both is expected to extract spatial-temporal representations for better decision-making. In the following, we describe the detailed architectures.
The proposed Vanilla\_TIT and Enhanced\_TIT backbones mainly consist of two Transformers as shown in Figure \ref{fig:TIT_First} and \ref{fig:TIT_Second}, respectively. The inner one is used to process a single observation; the outer one is responsible for processing the observation history; combining both is expected to extract spatial-temporal representations for better decision-making. %In the following, we describe the detailed architectures.

\subsection{The Vanilla\_TIT}\label{sec:VanillaTIT}
\vskip -0.06in
In this section, we aim at exploring \emph{the minimal implementation} of TIT. Therefore, compared to previous methods like GTrXL and CoBERL, the Vanilla\_TIT is designed by simply replacing their ResNet in the lower layers with the most basic Vision Transformer \cite{dosovitskiy2020image}, and further replacing their Transformer-XL or BERT with the most basic Transformer \cite{vaswani2017attention}. %We leave the exploration of more advanced Transformers like Swin Transformer \cite{liu2021swin} and GPT3 \cite{brown2020language} for the future work.

\subsubsection{The Inner Transformer}\label{sec:VanillaInnerTransformer}
The Inner Transformer operates on the \emph{observation patch}, which is similar to ViT \cite{dosovitskiy2020image}. To make our paper self-contained, we briefly introduce it with the following 6 steps. 

\textbf{(1) Observation Patch Generation.} 
Given an image observation $o \in \mathbb{R}^{H \times W \times C}$, we split it into a sequence of \textit{observation patches} $o^p \in \mathbb{R}^{N \times (P^2 \times C)}$, where $(H, W)$ is the resolution of the original image observation; $C$ is the number of channels; $(P, P)$ is the resolution of each patch; and $N = HW/P^2$ is the resulting number of patches, which is known as the context length for the Inner Transformer. 

If the observation is a 1-dimensional array with $D$ entries, i.e., $o \in \mathbb{R}^{D}$, we will take each entry as one patch independently. That is to say, the sequence of observation patches will be $o^p \in \mathbb{R}^{N \times 1}$ where $N=D$. This is convenient and reasonable because each entry usually has atomic semantics. For example, in the classic Mountain Car environment \footnote{\url{https://www.gymlibrary.dev/environments/classic_control/mountain_car/}}, the 2 entries of observation array represent atomic meaning of `Car position' and `Cart velocity', respectively. If there is prior knowledge, we can generate the observation patches in other ways (e.g., several entries as a patch). 

\textbf{(2) Observation Patch Embedding.} 
We map each observation patch $o^p_i$ into the \textit{observation patch embedding} with a trainable linear projection $E^p$:
\begin{eqnarray}
\hat{z}_0 = [o^p_1 E^p; o^p_2 E^p; ...; o^p_N E^p] \in \mathbb{R}^{N \times D^p}
\end{eqnarray}
where $E^p \in \mathbb{R}^{(P^2C) \times D^p}$ for image observations and $E^p \in \mathbb{R}^{1 \times D^p}$ for array observations, and $D^p$ is the dimension of the observation patch embedding.

\textbf{(3) Class Token Encoding.}
Similar to BERT and ViT, we also prepend a trainable \textit{class token} $z_0^{class} \in \mathbb{R}^{1 \times D^p}$:
\begin{eqnarray}
\check{z}_0 = [z_0^{class}; \hat{z}_0] \in \mathbb{R}^{(N+1) \times D^p}
\end{eqnarray}

\textbf{(4) Observation Patch Position Encoding.} 
We add \textit{observation patch position encoding}, which is a trainable parameter $E^p_{pos}$, to retain positional information:
\begin{eqnarray}
z_0 = \check{z}_0 + E^p_{pos}  \;  \;  \;  \;  \;  \;  E^p_{pos}  \in \mathbb{R}^{(N+1) \times D^p}
\end{eqnarray}
The resulting $z_0$ serves as the input of the Inner Transformer.

\textbf{(5) Inner Transformer Block.} 
We use the Transformer \textit{encoder} (i.e., without masking of other observation patches) since all patches in an observation can be used for decision-making. The operations in the $l$-th block $T^{in}_l$ are:
\begin{eqnarray}
\tilde{z}_l &=& z_{l-1} + \text{MSA}(\text{LN}(z_{l-1})) \;\;\;\;\;\; l = 1, ..., L \\
z_l &=& z_{l-1} + \text{FFN}(\text{LN}(\tilde{z}_{l-1})) \;\;\;\;\;\;\; l = 1, ..., L
\end{eqnarray}
where MSA, LN and FFN stand for the multiheaded self-attention (MSA), layer normalization (LN) and feed-forward network (FFN) used in the original Transformer \cite{vaswani2017attention}. In this way, the output of each block $z_l$ builds the \emph{spatial relationships} among observation patches within single observation, by computing interactions between any two observation patches. For example, in the Pong observation shown in Figure \ref{fig:TIT_First}, the patch containing the `Paddle' is more related to the patch containing the `Ball', but interacts less with other background patches.

\textbf{(6) Inner Transformer.} 
We stack $L$ blocks to form the Inner Transformer, so we finally get $z_L \in \mathbb{R}^{(N+1) \times D^p}$. As BERT and ViT, the first element $z_L[0] \in \mathbb{R}^{1 \times D^p}$ (i.e., the class token $z_L^{class}$) serves as the integrated representation of all observation patches. 

\subsubsection{The Outer Transformer}
\textbf{(1) Input Generation.}
In Vanilla\_TIT, the input of the Outer Transformer is exactly the output of the Inner Transformer. Specifically, since we make decisions based on the history of  observations $\langle o_{t-(K-1)}, ..., o_{t-1}, o_{t} \rangle$ in POMDP, the input of the Outer Transformer will be the concatenation of all $z_L[0]$s across $K$ timesteps, and we formally represent it as $y_0 = \{z_L[0]_t\}_{t-(K-1)}^{t} \in \mathbb{R}^{K \times D^p}$. Here, $K$ is the context length of the Outer Transformer. 

Note that we make decisions based on the history of tuples of $\langle \hat{R}_t, o_t, a_t \rangle$ in the learning paradigm of Decision Transformer \cite{chen2021decision}. In this case, we will replace $o_t$ by $z_L[0]_t$ as shown in Figure \ref{fig:TIT_First}.

\textbf{(2) Outer Transformer Block.} 
We use the Transformer \textit{decoder} (i.e., with masking of future observations) since future observations will not be available for decision-making during inference. The operations in the $l$-th block $T^{out}_l$ are:
\begin{eqnarray}
\tilde{y}_l &=& y_{l-1} + \text{MSA}(\text{LN}(y_{l-1})) \;\;\;\;\;\; l = 1, ..., L \\
y_l &=& y_{l-1} + \text{FFN}(\text{LN}(\tilde{y}_{l-1})) \;\;\;\;\;\;\; l = 1, ..., L
\end{eqnarray}
In this way, the output of each block $y_l$ builds the \emph{temporal relationships} among multiple consecutive observations, which is known to be helpful for the POMDP setting.

\textbf{(3) Outer Transformer.} 
We stack $L$ blocks to form the Outer Transformer, so we finally get $y_L \in \mathbb{R}^{K \times D^p}$. Since there is masking in Outer Transformer, only the last element of $y_L$ can fully represent the information of all observations. Thus, we apply an FFN upon the last element $y_L[K-1] \in \mathbb{R}^{1 \times D^p}$, which corresponds to the current timestep $t$, to generate the final action $a_t$ as shown by Figure \ref{fig:TIT_First}. 

\subsection{The Enhanced\_TIT}\label{sec:EnhancedTIT}
As shown in Figure \ref{fig:TIT_Second}, the Enhanced\_TIT has two improvements compared to Vanilla\_TIT. First, it builds a TIT block (i.e., the grey rectangle in Figure \ref{fig:TIT_Second}), which is composed of both the inner block and the outer block. Therefore, \emph{both spatial and temporal information} can be fused in \emph{every} TIT blocks, which may learn better representations for decision-making. Second, in order to stabilize training, we propose a dense connection design (i.e., the red dotted lines in Figure \ref{fig:TIT_Second}) to direct all immediate outputs of every TIT blocks to the final output explicitly. The effectiveness of this design has been verified by both CV methods like DenseNet \cite{Huang2017Densely} and RL network design studies like OFENet \cite{Ota2020Can} and D2RL\cite{sinha2020d2rl}.

\textbf{(1-4) Input Generation.} 
The Enhanced\_TIT also operates on the observation patch, so the following steps are the same as Vanilla\_TIT: (1) observation path generation; (2) observation patch embedding; (3) class token encoding; (4) observation patch position encoding. After these steps, we get $z_0 \in \mathbb{R}^{(N+1) \times D^p}$ as the input of the TIT block.

\textbf{(5) TIT Block.} 
The operations in the $l$-th TIT block are:
\begin{eqnarray}
z_l &=& T^{in}_l (z_{l-1}) \;\;\;\;\;\;\;\;\;\;\;\;\;\;\;\;\;\;\;\;\;\;\;\; l = 1, ..., L \\
y_l &=& T^{out}_l (\{z_l[0]_t\}_{t-(K-1)}^{t}) \;\;\;\;\;\;\; l = 1, ..., L
\end{eqnarray}
where $\{z_l[0]_t\}_{t-(K-1)}^{t}$ represents the concatenation of all $z_l[0]$s (i.e., all class token $z_l^{class}$s) across $K$ timesteps. Therefore, the spatial-temporal information is fused in every TIT blocks, which may learn more suitable representations for better decision-making.

\textbf{(6) TIT Backbone.}
We stack $L$ TIT blocks to form the Enhanced\_TIT backbone, so we finally get $\langle y_1, y_2, ..., y_L \rangle$, and each of them has a dimension of $y_l \in \mathbb{R}^{K \times D^p}$. We apply a dense connection to concatenate all of the last element $y_l[K-1] \in \mathbb{R}^{1 \times D^p}$, i.e., $y = \{y_l[K-1]\}_{l=1}^{L} \in \mathbb{R}^{L \times D^p}$, and apply an FFN upon the concatenation $y$ to generate the final action $a_t$ as shown by Figure \ref{fig:TIT_Second}.

\section{Experiment}
Our experiments focus on the following research questions (RQ). \textbf{RQ1:} Can TIT work better than other network architectures in the online RL setting? \textbf{RQ2:} Can TIT also work well in the offline setting? \textbf{RQ3:} Is there any deeper understanding of TIT by ablation and other approaches?

\subsection{Setting}
\textbf{Library Selection.} We have argued that TIT needs fewer optimization skills, and it can be used off-the-shelf by combining with popular RL libraries to achieve good performance. To verify this, we choose Stable-baseline3 (SB3) \cite{SB3} and d3rlpy \cite{d3rlpy} for our base implementation. The two libraries are widely-accepted by the community, and they ensure that the RL algorithms are implemented correctly. Besides, the two libraries provide credible benchmark results for popular environments to ensure a fair comparison. \emph{Unless otherwise noted, we only use the RL loss provided in these libraries to train TIT without applying any other optimization skills.}

\textbf{Algorithm Selection.} We also argued that TIT is agnostic to the RL training algorithms. In fact, it is general enough for most RL settings, e.g., both the online and offline settings, both on-policy and off-policy algorithms, both the Q-learning and policy gradient algorithms, environments with both the image observation and the array observation, and environments with both the discrete and continuous action spaces. However, there are too many combinations to test all. We decide to evaluate TIT with the PPO \cite{John2017PPO} algorithm from SB3 and the CQL \cite{kumar2020conservative} algorithm from d3rlpy, because they can cover a lot of settings as shown by the following table.
\begin{table}[h]
%\caption{PPO and CQL cover a lot of RL settings.}
\label{tab:Coverage_PPO_CQL}
\vskip -0.05in
\begin{center}
\begin{small}
\begin{sc}
\begin{tabular}{lcc}
\toprule
Setting & PPO & CQL \\
\midrule
Online / Offline & $\surd$ / $\times$ & $\times$ / $\surd$ \\
On-Policy / Off-Policy & $\surd$ / $\times$ & $\times$ / $\surd$ \\
Policy Gradient / Q-Learning & $\surd$ / $\times$ & $\times$ / $\surd$ \\
Image / Array Observation & $\surd$ / $\surd$ & $\surd$ / $\surd$ \\
Discrete / Continuous Action & $\surd$ / $\surd$ & $\surd$ / $\surd$ \\
\bottomrule
\end{tabular}
\end{sc}
\end{small}
\end{center}
\vskip -0.15in
\end{table}

To further verify the general applicability of TIT, we apply TIT to the offline supervised learning (SL) setting proposed by Decision Transformer (DT) \cite{chen2021decision}, which is a popular paradigm to solve long-term decision-making problems as offline RL does. We refer the readers to their original papers for more details of PPO, CQL and DT.

\textbf{Environment Selection.} Gym environments are typically used for evaluating new RL methods \cite{Gym2016Greg}, and we follow this practice. Specifically, for environments with image observations, we evaluate a few Gym Atari tasks like PongNoFrameskip-v4. For environments with array observations, the Gym Classic Control tasks like CartPole-v1 and the Gym MuJoCo tasks like Hopper-v3 are used for evaluation. We choose these tasks because they are popularly used, and also have been credibly benchmarked by SB3, d3rlpy and DT. %Specifically, for environments with array observations, the Acrobot-v1, CartPole-v1 and MountainCar-v0 from Gym Classic Control, and the Ant-v3, Hopper-v3 and Walker2d-v3 from Gym MuJoCo are used for evaluation. For environments with image observations, we evaluate the BreakoutNoFrameskip-v4, MsPacmanNoFrameskip-v4, PongNoFrameskip-v4 and SpaceInvadersNoFrameskip-v4 from Gym Atari.

\textbf{Baseline Network Architecture Selection.} For environments with image observations, we compare TIT with NatureCNN + MLP \cite{mnih2015human}, ResNet + MLP \cite{RRL2021Shah}, ResNet + Transformer + MLP (i.e., Catformer \cite{davis2021catformer}), ResNet + Transformer + Gating + LSTM + MLP (i.e., CoBERL \cite{banino2021coberl}). For environments with array observations, we compare TIT with the Pure MLP, OFENet \cite{Ota2020Can} and D2RL \cite{sinha2020d2rl}. As mentioned before, OFENet and D2RL design larger and deeper networks by applying different variants of DenseNet. Similarly, Catformer takes the concatenation of the outputs of all previous layers as the input of the current layer to reduce sensitivity. The dense connection in the output layer of Enhanced\_TIT adopts a similar design. So we can roughly know whether the Transformers are really helpful by comparing with these architectures.

\textbf{Hyperparameter Selection.} We implement the above network architectures as closely as possible to their original papers and open-source code repositories. \emph{Except for the network architectures, we do not change any hyperparameters and settings of SB3, d3rlpy and DT for fair comparison.} The detailed settings can be found in Appendix \ref{sec:ExperimentDetails_Appendix}.

\subsection{RQ1: Network Architecture Comparison}
The results for tasks with image-based and array-based observations are shown in Table \ref{tab:OnlineResultsImageObs} and \ref{tab:OnlineResultsArrayObs}, respectively. The following conclusions can be drawn from these results. 

\textbf{(1) Our reproduction is correct.} In these tables, `Reported' means the results reported by official SB3 (see Appendix \ref{sec:ReferencedResults_Appendix} for details), while the `NatureCNN + MLP (Repro)' and `Pure MLP (Repro)' represent our reproduction of PPO based on specific network architectures. As we can see, NatureCNN + MLP and Pure MLP achieve similar or better results than Reported in nine out of ten tasks, and the only exception is MountainCar, which is known to be sensitive to solve. These results indicate that our code-level implementation of PPO is correct, so we can ensure a credible comparison based on our implementation.

\textbf{(2) Our motivation has been verified.} The results in Table \ref{tab:OnlineResultsImageObs} demonstrate that Catformer and CoBERL are unstable across different tasks. For example, they get almost the full score (i.e., 21.0) in Pong, but get low scores in other tasks. When we tune them carefully with many optimization skills proposed in the papers (we call the resulting algorithms Catformer+tuned and CoBERL+tuned), they achieve higher scores than before as shown by the last two rows of Table \ref{tab:OnlineResultsImageObs}, but they still cannot get superior performance compared to NatureCNN + MLP, ResNet + MLP and our Enhanced\_TIT. This shows that previous Transformer-based deep RL methods, which are usually composed of several mixed modules, are hard to optimize.

\textbf{(3) Enhanced\_TIT achieves comparable performance than several strong baselines in different tasks.} Specifically, it achieves almost full scores in Pong and CartPole, the highest scores in MsPacman and MountainCar, and comparable scores to the best baselines in other Atari and Classic Control tasks. However, Enhanced\_TIT performs slightly worse than the state-of-the-arts in MoJoCo tasks, although it has obtained good scores. In summary, these results demonstrate the potential of pure-Transformer networks for online RL settings. % We are currently unable to figure out the exact reasons, but hypothesize that 1) the action space is an important factor for pure Transformer-based RL methods because Classic Control and Atari have discrete action space while MoJoCo has continuous action space; 2) the inner Transformer is actually a ViT, which is good at processing image observation, but MuJoCo has array observation. We will investigate these hypotheses in the future. % ; 3) the MuJoCo tasks aim to increase the number of independent state that often has a value of zero, which may hinder the learning of our modeling method (recall that we take each entry of state as one patch as mentioned in Section \ref{sec:VanillaInnerTransformer}, so there will be many patch embeddings with zero values)

\begin{table*}[h]
\caption{The results for online Atari tasks. Compared to other Transformer-based networks, Enhanced\_TIT is the most stable across tasks.}
\label{tab:OnlineResultsImageObs}
\vskip 0.15in
\begin{center}
\begin{small}
\begin{sc}
\begin{tabular}{l|cccccccc}
\toprule
Task Name & \multicolumn{2}{c}{Breakout} & \multicolumn{2}{c}{MsPacman} & \multicolumn{2}{c}{Pong} & \multicolumn{2}{c}{SpaceInvaders} \\
Obs/Act Space & (1, 84, 84) & 4d & (1, 84, 84) & 9d & (1, 84, 84) & 9d & (1, 84, 84) & 6d \\
\midrule
Episode Return & mean & std & mean & std & mean & std & mean & std \\
\midrule
Reported & \cellcolor{gray!20}\textbf{398} & 33 & 1754 & 172 & \cellcolor{gray!20}\textbf{20.989} & 0.105 & 960 & 425 \\
NatureCNN+MLP (Repro) & \cellcolor{gray!20}\textbf{391} & 26 & \cellcolor{gray!20}\textbf{2111} & 589 & \cellcolor{gray!20}\textbf{21.000} & 0.000 & \cellcolor{gray!20}\textbf{1455} & 387 \\
\midrule
\cellcolor{red!10}\textbf{Vanilla\_TIT (ours)} & 169 & 91 & 748 & 205 & 9.600 & 6.445 & 752 & 77 \\
\cellcolor{red!10}\textbf{Enhanced\_TIT (ours)} & 321 & 68 & \cellcolor{gray!20}\textbf{2246} & 326 & \cellcolor{gray!20}\textbf{20.750} & 1.577 & \cellcolor{gray!20}\textbf{1645} & 168 \\
\midrule
ResNet+MLP & \cellcolor{gray!20}\textbf{397} & 57 & 1807 & 405 & \cellcolor{gray!20}\textbf{21.000} & 0.000  & \cellcolor{gray!20}\textbf{1700} & 511 \\
Catformer & 165 & 57 & 427 & 388 & \cellcolor{gray!20}\textbf{19.980} & 0.139 & \cellcolor{gray!20}\textbf{1427} & 597 \\
CoBERL & 189 & 26 & 195 & 252 & \cellcolor{gray!20}\textbf{19.460} & 1.557 & 618 & 245 \\
\midrule
Catformer+tuned & 242 & 41 & 1579 & 461 & - & - & - & - \\
CoBERL+tuned & 358 & 34 & \cellcolor{gray!20}\textbf{2190} & 327 & - & - & 821 & 314 \\
\bottomrule
\end{tabular}
\end{sc}
\end{small}
\end{center}
\vskip -0.1in
\end{table*}

\begin{table*}[h]
\caption{The results for online Classic Control and MoJoCo tasks. In the table, `OBS/ACT Space' means observation and action space, and 3D/8C means the action space is discrete/continuous with 3/8 entries; and there are diverse settings for `OBS/ACT Space'.}
\label{tab:OnlineResultsArrayObs}
\vskip 0.15in
\begin{center}
\begin{small}
\begin{sc}
\begin{tabular}{l|cccccc|cccccc}
\toprule
Task Type & \multicolumn{6}{c|}{Classic Control} & \multicolumn{6}{c}{MoJoCo} \\
Task Name & \multicolumn{2}{c}{Acrobot} & \multicolumn{2}{c}{CartPole} & \multicolumn{2}{c|}{MountainCar} & \multicolumn{2}{c}{Ant} & \multicolumn{2}{c}{Hopper} & \multicolumn{2}{c}{Walker2d} \\
Obs/Act Space & 6 & 3d & 4 & 2d & 2 & 3d & 111 & 8c & 11 & 3c & 17 & 6c \\
\midrule
Episode Return & mean & std & mean & std & mean & std & mean & std & mean & std & mean & std \\
\midrule
Reported & \cellcolor{gray!20}\textbf{-73} & 18 & \cellcolor{gray!20}\textbf{500} & 0 & -110 & 19 & 1327 & 451 & 2410 & 10 & \cellcolor{gray!20}\textbf{3478} & 821 \\
Pure MLP (Repro) & -81 & 11 & \cellcolor{gray!20}\textbf{500} & 0 & -200 & 0 & \cellcolor{gray!20}\textbf{2284} & 584 & \cellcolor{gray!20}\textbf{3530} & 13 & \cellcolor{gray!20}\textbf{3306} & 1119 \\
\midrule
\cellcolor{red!10}\textbf{Vanilla\_TIT} & -84 & 19 & \cellcolor{gray!20}\textbf{500} & 0 & \cellcolor{gray!20}\textbf{-96} & 7 & 976 & 186 & 1060 & 56 & 1701 & 323 \\
\cellcolor{red!10}\textbf{Enhanced\_TIT} & -81 & 17 & \cellcolor{gray!20}\textbf{500} & 0 & \cellcolor{gray!20}\textbf{-97} & 7 & 1975 & 234 & 2008 & 48 & 2911 & 249 \\
\midrule
OFENet & -81 & 10 & \cellcolor{gray!20}\textbf{500} & 0 & -200 & 0 & 2111 & 542 & \cellcolor{gray!20}\textbf{3483} & 17 & \cellcolor{gray!20}\textbf{3956} & 611 \\
D2RL & -83 & 11 & \cellcolor{gray!20}\textbf{500} & 0 & -200 & 0 & 1825 & 875 & \cellcolor{gray!20}\textbf{3565} & 65 & 2141 & 774 \\
\bottomrule
\end{tabular}
\end{sc}
\end{small}
\end{center}
\vskip -0.1in
\end{table*}

\begin{table*}[t!]
\caption{The normalized scores for offline MoJoCo datasets from D4RL \cite{fu2020d4rl}. According to the normalization rule shown in Appendix \ref{sec:CQL_TIT_Hyperparameter}, a small increase of the normalized score means a large improvement of the original score. Note that CQL and DT use different versions of offline datasets (see Appendix \ref{sec:Dataset_Appendix} for the details), so the results of CQL and DT may not be comparable; but for each separate algorithm, our CQL\_TIT is superior to CQL, and our DT\_TIT is superior to DT.}
\label{tab:OfflineMain}
\vskip 0.15in
\begin{center}
\begin{small}
\begin{sc}
\begin{tabular}{ll|ccc|ccc}
\toprule
Dataset & Task Name & CQL\_Reported & Repro & \cellcolor{red!10}\textbf{CQL\_TIT} & DT\_Reported & Repro & \cellcolor{red!10}\textbf{DT\_TIT} \\
\midrule
\multirow{3}{*}{Medium} & Halfcheetah & \cellcolor{gray!20}\textbf{42.6$\pm$0.1} & \cellcolor{gray!20}\textbf{42.6$\pm$0.1} & \cellcolor{gray!20}\textbf{42.6$\pm$0.1} & \cellcolor{gray!20}\textbf{42.6$\pm$0.1} & \cellcolor{gray!20}\textbf{42.5$\pm$2.3} & \cellcolor{gray!20}\textbf{42.8$\pm$2.3} \\
& Hopper & \cellcolor{gray!20}\textbf{100.7$\pm$0.3} & \cellcolor{gray!20}\textbf{100.8$\pm$0.3} & \cellcolor{gray!20}\textbf{100.8$\pm$0.2} & 67.6$\pm$1.0 & 67.4$\pm$4.1 & \cellcolor{gray!20}\textbf{68.2$\pm$2.4} \\
& Walker2d & 82.8$\pm$1.3 & 83.1$\pm$1.2 & \cellcolor{gray!20}\textbf{84.1$\pm$0.9} & 74.0$\pm$1.4 & 75.2$\pm$0.6 & \cellcolor{gray!20}\textbf{77.6$\pm$0.6} \\
\midrule
\multirow{3}{*}{\makecell{Medium \\ Replay}} & Halfcheetah & 47.1$\pm$0.6 & 46.1$\pm$0.1 & \cellcolor{gray!20}\textbf{47.8$\pm$0.1} & 36.6$\pm$0.8 & 37.0$\pm$2.8 & \cellcolor{gray!20}\textbf{40.8$\pm$2.3} \\
& Hopper & 85.1$\pm$16.2 & 58.8$\pm$12.7 & \cellcolor{gray!20}\textbf{99.2$\pm$1.7} & 82.7$\pm$7.0 & 71.0$\pm$4.7 & \cellcolor{gray!20}\textbf{89.6$\pm$2.7} \\
& Walker2d & 49.6$\pm$5.2 & 49.5$\pm$2.9 & \cellcolor{gray!20}\textbf{53.6$\pm$3.7} & 66.6$\pm$3.0 & 71.0$\pm$0.5 & \cellcolor{gray!20}\textbf{74.1$\pm$0.6} \\
\bottomrule
\end{tabular}
\end{sc}
\end{small}
\end{center}
\vskip -0.1in
\end{table*}

\subsection{RQ2: Applying TIT in Offline Setting}
As mentioned before, we apply TIT to both the offline RL setting based on CQL and the offline SL setting based on DT. We call our methods CQL\_TIT and DT\_TIT, respectively. Table \ref{tab:OfflineMain} shows the normalized scores, which demonstrate the following conclusions.

\textbf{(1) Our implementation of CQL and DT is correct.} Specifically, `CQL\_Reported' and `DT\_Reported' represent the official results (see Appendix \ref{sec:ReferencedResults_Appendix} for details), while `Repro' means our reproduced results, and they have very close values in most datasets. Therefore, we can ensure a credible comparison based on our correct implementation.

\textbf{(2) For the offline RL setting, CQL\_TIT is better than CQL, especially in practical datasets with complex distributions.} Specifically, CQL\_TIT matches or exceeds CQL by a small margin in `Medium' datasets that are generated from a single policy. Furthermore, in `Medium Replay' datasets that combine multiple policies \footnote{The `Medium Replay' is used in the original DT paper, while it is called `Mixed' dataset in the original CQL paper.}, CQL\_TIT outperforms CQL by a large margin. Since the mixed datasets with complex distributions are more likely to be common in practice \cite{kumar2020conservative}, we expect that CQL\_TIT will work better than CQL in practical applications.

\textbf{(3) For the offline SL setting, DT\_TIT is better than DT, especially in practical datasets.} The analysis is similar to that of the offline RL setting.

\textbf{(4) The offline TIT can achieve better results than the online TIT in MoJoCo tasks.} This is easy to see by comparing Table \ref{tab:OnlineResultsArrayObs} and \ref{tab:OfflineMain}. We discuss some possible reasons for this in Appendix \ref{appendix:Discussion}.

\subsection{RQ3: Deeper Understanding of TIT}
Here we show the ablation study and feature visualization. More results and analyses are shown in the Appendix \ref{appendix:MoreAnalyses}.

\subsubsection{Ablation Study}
Vanilla\_TIT can be regarded as an ablation model of Enhanced\_TIT. Table \ref{tab:OnlineResultsImageObs} and \ref{tab:OnlineResultsArrayObs} show that although Vanilla\_TIT can achieve good results in simple Classic Control tasks, its performance is worse than Enhanced\_TIT in other tasks.

In this section, we further consider three ablation models: `w/o Dense' means that the dense connection is removed; `w/o Inner' means that the inner blocks are removed; `w/o Outer' means that the outer blocks are removed. We show some results in Table \ref{tab:AblationMain}. The details of these models and more results can be found in Appendix \ref{appendix:Ablation}. 

Compared to Enhanced\_TIT, `w/o Dense' has the maximum performance degradation. \textbf{It implies that the dense connection is important for pure Transformer-based networks}. As far as we know, previous studies have shown that the dense connection is also important for the CNN \cite{Huang2017Densely} and MLP \cite{Ota2020Can,sinha2020d2rl} networks, but TIT is the first work to demonstrate this for Transformer networks. 

Furthermore, both `w/o Inner' and `w/o Outer' perform worse than Enhanced\_TIT, which indicates that both inner and outer blocks (and proper arrangement of them) are necessary for good performance.

\begin{table}[t]
\caption{The results of ablation models in Atari tasks.}
\label{tab:AblationMain}
\vskip 0.15in
\begin{center}
\begin{small}
\begin{sc}
\begin{tabular}{l|cccccccc}
\toprule
Task Name & \multicolumn{2}{c}{Pong} & \multicolumn{2}{c}{SpaceInvaders} \\
\midrule
Episode Return & mean & std & mean & std \\
\midrule
Vanilla\_TIT & 9.600 & 6.445 & 752 & 77 \\
Enhanced\_TIT & 20.750 & 1.577 & 1645 & 168 \\
\midrule
w/o Dense & 18.620 & 3.267 & 938 & 256 \\
w/o Inner & 19.350 & 2.441 & 1363 & 91 \\
w/o Outer & 20.180 & 2.034 & 1295 & 233 \\
\bottomrule
\end{tabular}
\end{sc}
\end{small}
\end{center}
\vskip -0.1in
\end{table}

\subsubsection{Feature Visualization}
\begin{figure}[t]
\vskip 0.1in
\begin{center}
\centerline{\includegraphics[width=0.92\columnwidth]{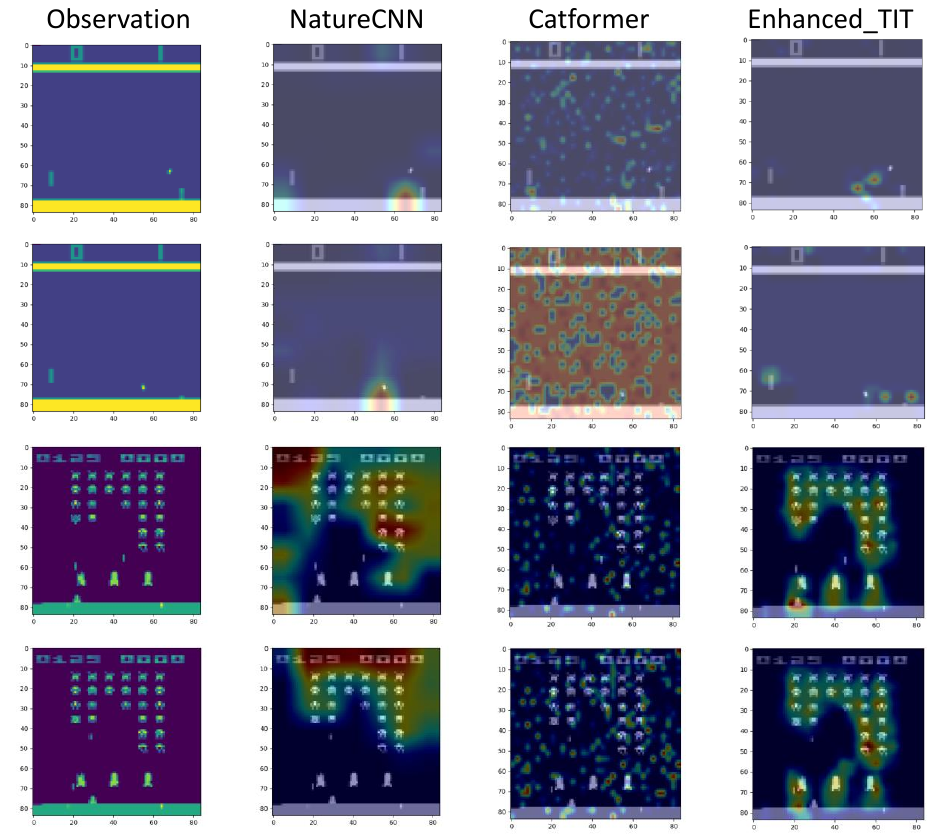}}
\caption{The feature attention of different methods. Observations (i.e., the first column) are randomly generated from Pong (i.e., the first two rows) and SpaceInvaders (i.e., the last two rows).}
\label{fig:GradCAM_CompareMain}
\end{center}
\vskip -0.3in
\end{figure}

We visualize the feature attention of different methods in Figure \ref{fig:GradCAM_CompareMain}, which presents the following phenomena.

\textbf{(1) From the spatial perspective, Enhanced\_TIT has more explainable attention maps than other methods.} Specifically, its attention is highly correlated with the objects in the original observation in most cases (only the first row is an exception). Furthermore, in the SpaceInvaders task, the invaders close to the agent get more attention (i.e., the visualization color is darker). In this way, Enhanced\_TIT may learn better spatial representations for good decision-making. In contrast, Catformer has disorganized attention maps, and it sometimes generates unexplainable attention as shown by the second row. NatureCNN has organized attention maps, but it is not so accurate at the object level.

\textbf{(2) From the temporal perspective, Enhanced\_TIT has more stable attention maps than other methods.} For example, when the observation changes slightly (e.g., the second row is slightly changed compared to the first; so does the fourth compared to the third), its attention map does not change significantly. Therefore, Enhanced\_TIT may learn consistent temporal representations for good decision-making and its stable performance. In contrast, the attention map of NatureCNN and Catformer has changed a lot. % which may explain its stable performance

\section{Conclusion}
This paper explored the minimal implementation of Transformer in Transformer (TIT) backbone for deep RL. The key idea is cascading two Transformers in a very natural way: the inner one is used to process a single observation, while the outer one is responsible for processing the history of observations; combining both can extract spatial-temporal representations for better decision-making. The experiments demonstrated that TIT can be used off-the-shelf as backbones to achieve good results without tedious optimization skills and the combination of other network modules. The experiments also presented deeper understanding of TIT. The limitations and future directions are discussed in Appendix \ref{appendix:Discussion}.

\bibliography{example_paper}

\begin{thebibliography}{59}
\providecommand{\natexlab}[1]{#1}
\providecommand{\url}[1]{\texttt{#1}}
\expandafter\ifx\csname urlstyle\endcsname\relax
  \providecommand{\doi}[1]{doi: #1}\else
  \providecommand{\doi}{doi: \begingroup \urlstyle{rm}\Url}\fi

\bibitem[Arnab et~al.(2021)Arnab, Dehghani, Heigold, Sun, Lučić, and
  Schmid]{Arnab2021ViViT}
Arnab, A., Dehghani, M., Heigold, G., Sun, C., Lučić, M., and Schmid, C.
\newblock Vivit: A video vision transformer.
\newblock In \emph{2021 IEEE/CVF International Conference on Computer Vision
  (ICCV)}, pp.\  6816--6826, 2021.
\newblock \doi{10.1109/ICCV48922.2021.00676}.

\bibitem[Babaeizadeh et~al.(2016)Babaeizadeh, Frosio, Tyree, Clemons, and
  Kautz]{babaeizadeh2016ga3c}
Babaeizadeh, M., Frosio, I., Tyree, S., Clemons, J., and Kautz, J.
\newblock Ga3c: Gpu-based a3c for deep reinforcement learning.
\newblock \emph{CoRR abs/1611.06256}, 2016.

\bibitem[Banino et~al.(2022)Banino, Badia, Walker, Scholtes, Mitrovic, and
  Blundell]{banino2021coberl}
Banino, A., Badia, A.~P., Walker, J.~C., Scholtes, T., Mitrovic, J., and
  Blundell, C.
\newblock Coberl: Contrastive bert for reinforcement learning.
\newblock In \emph{International Conference on Learning Representations}, 2022.

\bibitem[Beattie et~al.(2016)Beattie, Leibo, Teplyashin, Ward, Wainwright,
  K{\"u}ttler, Lefrancq, Green, Vald{\'e}s, Sadik, et~al.]{beattie2016deepmind}
Beattie, C., Leibo, J.~Z., Teplyashin, D., Ward, T., Wainwright, M.,
  K{\"u}ttler, H., Lefrancq, A., Green, S., Vald{\'e}s, V., Sadik, A., et~al.
\newblock Deepmind lab.
\newblock \emph{arXiv preprint arXiv:1612.03801}, 2016.

\bibitem[Bellemare et~al.(2017)Bellemare, Dabney, and
  Munos]{bellemare2017distributional}
Bellemare, M.~G., Dabney, W., and Munos, R.
\newblock A distributional perspective on reinforcement learning.
\newblock In \emph{International Conference on Machine Learning}, pp.\
  449--458. PMLR, 2017.

\bibitem[Bjorck et~al.(2021)Bjorck, Gomes, and Weinberger]{bjorck2021towards}
Bjorck, N., Gomes, C.~P., and Weinberger, K.~Q.
\newblock Towards deeper deep reinforcement learning with spectral
  normalization.
\newblock \emph{Advances in Neural Information Processing Systems},
  34:\penalty0 8242--8255, 2021.

\bibitem[Brockman et~al.(2016)Brockman, Cheung, Pettersson, Schneider,
  Schulman, Tang, and Zaremba]{Gym2016Greg}
Brockman, G., Cheung, V., Pettersson, L., Schneider, J., Schulman, J., Tang,
  J., and Zaremba, W.
\newblock Openai gym, 2016.

\bibitem[Brown et~al.(2020)Brown, Mann, Ryder, Subbiah, Kaplan, Dhariwal,
  Neelakantan, Shyam, Sastry, Askell, et~al.]{brown2020language}
Brown, T., Mann, B., Ryder, N., Subbiah, M., Kaplan, J.~D., Dhariwal, P.,
  Neelakantan, A., Shyam, P., Sastry, G., Askell, A., et~al.
\newblock Language models are few-shot learners.
\newblock \emph{Advances in neural information processing systems},
  33:\penalty0 1877--1901, 2020.

\bibitem[Chen et~al.(2021{\natexlab{a}})Chen, Yoon, Wu, and
  Ahn]{chen2021transdreamer}
Chen, C., Yoon, J., Wu, Y.-F., and Ahn, S.
\newblock Transdreamer: Reinforcement learning with transformer world models.
\newblock In \emph{Deep RL Workshop NeurIPS 2021}, 2021{\natexlab{a}}.

\bibitem[Chen et~al.(2021{\natexlab{b}})Chen, Lu, Rajeswaran, Lee, Grover,
  Laskin, Abbeel, Srinivas, and Mordatch]{chen2021decision}
Chen, L., Lu, K., Rajeswaran, A., Lee, K., Grover, A., Laskin, M., Abbeel, P.,
  Srinivas, A., and Mordatch, I.
\newblock Decision transformer: Reinforcement learning via sequence modeling.
\newblock \emph{Advances in neural information processing systems},
  34:\penalty0 15084--15097, 2021{\natexlab{b}}.

\bibitem[Dabney et~al.(2018)Dabney, Rowland, Bellemare, and
  Munos]{dabney2018distributional}
Dabney, W., Rowland, M., Bellemare, M., and Munos, R.
\newblock Distributional reinforcement learning with quantile regression.
\newblock In \emph{Proceedings of the AAAI Conference on Artificial
  Intelligence}, volume~32, 2018.

\bibitem[Dai et~al.(2019)Dai, Yang, Yang, Carbonell, Le, and
  Salakhutdinov]{dai2019transformer}
Dai, Z., Yang, Z., Yang, Y., Carbonell, J.~G., Le, Q., and Salakhutdinov, R.
\newblock Transformer-xl: Attentive language models beyond a fixed-length
  context.
\newblock In \emph{Proceedings of the 57th Annual Meeting of the Association
  for Computational Linguistics}, pp.\  2978--2988, 2019.

\bibitem[Davis et~al.(2021)Davis, Gu, Choromanski, Dao, Re, Finn, and
  Liang]{davis2021catformer}
Davis, J.~Q., Gu, A., Choromanski, K., Dao, T., Re, C., Finn, C., and Liang, P.
\newblock Catformer: Designing stable transformers via sensitivity analysis.
\newblock In \emph{International Conference on Machine Learning}, pp.\
  2489--2499. PMLR, 2021.

\bibitem[Dosovitskiy et~al.(2020)Dosovitskiy, Beyer, Kolesnikov, Weissenborn,
  Zhai, Unterthiner, Dehghani, Minderer, Heigold, Gelly,
  et~al.]{dosovitskiy2020image}
Dosovitskiy, A., Beyer, L., Kolesnikov, A., Weissenborn, D., Zhai, X.,
  Unterthiner, T., Dehghani, M., Minderer, M., Heigold, G., Gelly, S., et~al.
\newblock An image is worth 16x16 words: Transformers for image recognition at
  scale.
\newblock In \emph{International Conference on Learning Representations}, 2020.

\bibitem[Evans \& Gao(2016)Evans and Gao]{evans2016deepmind}
Evans, R. and Gao, J.
\newblock Deepmind ai reduces google data centre cooling bill by 40\%.
\newblock \emph{DeepMind blog}, 20:\penalty0 158, 2016.

\bibitem[Fu et~al.(2020)Fu, Kumar, Nachum, Tucker, and Levine]{fu2020d4rl}
Fu, J., Kumar, A., Nachum, O., Tucker, G., and Levine, S.
\newblock D4rl: Datasets for deep data-driven reinforcement learning, 2020.

\bibitem[Fujimoto et~al.(2018)Fujimoto, Hoof, and
  Meger]{fujimoto2018addressing}
Fujimoto, S., Hoof, H., and Meger, D.
\newblock Addressing function approximation error in actor-critic methods.
\newblock In \emph{International conference on machine learning}, pp.\
  1587--1596. PMLR, 2018.

\bibitem[Fujimoto et~al.(2019)Fujimoto, Meger, and Precup]{fujimoto2019off}
Fujimoto, S., Meger, D., and Precup, D.
\newblock Off-policy deep reinforcement learning without exploration.
\newblock In \emph{International conference on machine learning}, pp.\
  2052--2062. PMLR, 2019.

\bibitem[Gildenblat \& contributors(2021)Gildenblat and contributors]{GradCAM}
Gildenblat, J. and contributors.
\newblock Pytorch library for cam methods.
\newblock \url{https://github.com/jacobgil/pytorch-grad-cam}, 2021.

\bibitem[Goul{\~a}o \& Oliveira(2022)Goul{\~a}o and
  Oliveira]{goulao2022pretraining}
Goul{\~a}o, M. and Oliveira, A.~L.
\newblock Pretraining the vision transformer using self-supervised methods for
  vision based deep reinforcement learning.
\newblock \emph{arXiv preprint arXiv:2209.10901}, 2022.

\bibitem[Greff et~al.(2016)Greff, Srivastava, Koutn{\'\i}k, Steunebrink, and
  Schmidhuber]{greff2016lstm}
Greff, K., Srivastava, R.~K., Koutn{\'\i}k, J., Steunebrink, B.~R., and
  Schmidhuber, J.
\newblock Lstm: A search space odyssey.
\newblock \emph{IEEE transactions on neural networks and learning systems},
  28\penalty0 (10):\penalty0 2222--2232, 2016.

\bibitem[Hafner et~al.(2019)Hafner, Lillicrap, Ba, and
  Norouzi]{hafner2019dream}
Hafner, D., Lillicrap, T., Ba, J., and Norouzi, M.
\newblock Dream to control: Learning behaviors by latent imagination.
\newblock \emph{arXiv preprint arXiv:1912.01603}, 2019.

\bibitem[Han et~al.(2021)Han, Xiao, Wu, Guo, Xu, and Wang]{han2021transformer}
Han, K., Xiao, A., Wu, E., Guo, J., Xu, C., and Wang, Y.
\newblock Transformer in transformer.
\newblock \emph{Advances in Neural Information Processing Systems},
  34:\penalty0 15908--15919, 2021.

\bibitem[Hausknecht \& Stone(2015)Hausknecht and Stone]{hausknecht2015deep}
Hausknecht, M. and Stone, P.
\newblock Deep recurrent q-learning for partially observable mdps.
\newblock In \emph{2015 aaai fall symposium series}, 2015.

\bibitem[He et~al.(2016)He, Zhang, Ren, and Sun]{he2016deep}
He, K., Zhang, X., Ren, S., and Sun, J.
\newblock Deep residual learning for image recognition.
\newblock In \emph{Proceedings of the IEEE conference on computer vision and
  pattern recognition}, pp.\  770--778, 2016.

\bibitem[Hu et~al.(2022)Hu, Shen, Zhang, Chen, and Tao]{hu2022TRL}
Hu, S., Shen, L., Zhang, Y., Chen, Y., and Tao, D.
\newblock On transforming reinforcement learning by transformer: The
  development trajectory.
\newblock \emph{arXiv preprint arXiv:2212.14164}, 2022.

\bibitem[Huang et~al.(2017)Huang, Liu, Van Der~Maaten, and
  Weinberger]{Huang2017Densely}
Huang, G., Liu, Z., Van Der~Maaten, L., and Weinberger, K.~Q.
\newblock Densely connected convolutional networks.
\newblock In \emph{2017 IEEE Conference on Computer Vision and Pattern
  Recognition (CVPR)}, pp.\  2261--2269, 2017.
\newblock \doi{10.1109/CVPR.2017.243}.

\bibitem[Janner et~al.(2021)Janner, Li, and Levine]{janner2021reinforcement}
Janner, M., Li, Q., and Levine, S.
\newblock Reinforcement learning as one big sequence modeling problem.
\newblock In \emph{ICML 2021 Workshop on Unsupervised Reinforcement Learning},
  2021.

\bibitem[Kalashnikov et~al.(2018)Kalashnikov, Irpan, Pastor, Ibarz, Herzog,
  Jang, Quillen, Holly, Kalakrishnan, Vanhoucke,
  et~al.]{kalashnikov2018scalable}
Kalashnikov, D., Irpan, A., Pastor, P., Ibarz, J., Herzog, A., Jang, E.,
  Quillen, D., Holly, E., Kalakrishnan, M., Vanhoucke, V., et~al.
\newblock Scalable deep reinforcement learning for vision-based robotic
  manipulation.
\newblock In \emph{Conference on Robot Learning}, pp.\  651--673. PMLR, 2018.

\bibitem[Kenton \& Toutanova(2019)Kenton and Toutanova]{kenton2019bert}
Kenton, J. D. M.-W.~C. and Toutanova, L.~K.
\newblock Bert: Pre-training of deep bidirectional transformers for language
  understanding.
\newblock In \emph{Proceedings of NAACL-HLT}, pp.\  4171--4186, 2019.

\bibitem[Kumar et~al.(2020)Kumar, Zhou, Tucker, and
  Levine]{kumar2020conservative}
Kumar, A., Zhou, A., Tucker, G., and Levine, S.
\newblock Conservative q-learning for offline reinforcement learning.
\newblock \emph{Advances in Neural Information Processing Systems},
  33:\penalty0 1179--1191, 2020.

\bibitem[Liang et~al.(2022)Liang, Zhou, Zimmermann, and
  Yan]{Liang2022DualFormer}
Liang, Y., Zhou, P., Zimmermann, R., and Yan, S.
\newblock Dualformer: Local-global stratified transformer for efficient video
  recognition.
\newblock In Avidan, S., Brostow, G., Ciss{\'e}, M., Farinella, G.~M., and
  Hassner, T. (eds.), \emph{Computer Vision -- ECCV 2022}, pp.\  577--595,
  Cham, 2022. Springer Nature Switzerland.
\newblock ISBN 978-3-031-19830-4.

\bibitem[Lillicrap et~al.(2015)Lillicrap, Hunt, Pritzel, Heess, Erez, Tassa,
  Silver, and Wierstra]{lillicrap2015continuous}
Lillicrap, T.~P., Hunt, J.~J., Pritzel, A., Heess, N., Erez, T., Tassa, Y.,
  Silver, D., and Wierstra, D.
\newblock Continuous control with deep reinforcement learning.
\newblock \emph{arXiv preprint arXiv:1509.02971}, 2015.

\bibitem[Liu et~al.(2021)Liu, Lin, Cao, Hu, Wei, Zhang, Lin, and
  Guo]{liu2021swin}
Liu, Z., Lin, Y., Cao, Y., Hu, H., Wei, Y., Zhang, Z., Lin, S., and Guo, B.
\newblock Swin transformer: Hierarchical vision transformer using shifted
  windows.
\newblock In \emph{Proceedings of the IEEE/CVF International Conference on
  Computer Vision}, pp.\  10012--10022, 2021.

\bibitem[Melo(2022)]{melo2022transformers}
Melo, L.~C.
\newblock Transformers are meta-reinforcement learners.
\newblock In \emph{International Conference on Machine Learning}, pp.\
  15340--15359. PMLR, 2022.

\bibitem[Meng et~al.(2022)Meng, Goodwin, Yazidi, and Engelstad]{meng2022deep}
Meng, L., Goodwin, M., Yazidi, A., and Engelstad, P.
\newblock Deep reinforcement learning with swin transformer.
\newblock \emph{arXiv preprint arXiv:2206.15269}, 2022.

\bibitem[Micheli et~al.(2022)Micheli, Alonso, and
  Fleuret]{micheli2022transformers}
Micheli, V., Alonso, E., and Fleuret, F.
\newblock Transformers are sample efficient world models.
\newblock \emph{arXiv preprint arXiv:2209.00588}, 2022.

\bibitem[Mnih et~al.(2015)Mnih, Kavukcuoglu, Silver, Rusu, Veness, Bellemare,
  Graves, Riedmiller, Fidjeland, Ostrovski, et~al.]{mnih2015human}
Mnih, V., Kavukcuoglu, K., Silver, D., Rusu, A.~A., Veness, J., Bellemare,
  M.~G., Graves, A., Riedmiller, M., Fidjeland, A.~K., Ostrovski, G., et~al.
\newblock Human-level control through deep reinforcement learning.
\newblock \emph{nature}, 518\penalty0 (7540):\penalty0 529--533, 2015.

\bibitem[Mnih et~al.(2016)Mnih, Badia, Mirza, Graves, Lillicrap, Harley,
  Silver, and Kavukcuoglu]{mnih2016asynchronous}
Mnih, V., Badia, A.~P., Mirza, M., Graves, A., Lillicrap, T., Harley, T.,
  Silver, D., and Kavukcuoglu, K.
\newblock Asynchronous methods for deep reinforcement learning.
\newblock In \emph{International conference on machine learning}, pp.\
  1928--1937. PMLR, 2016.

\bibitem[Okada \& Taniguchi(2021)Okada and Taniguchi]{okada2021dreaming}
Okada, M. and Taniguchi, T.
\newblock Dreaming: Model-based reinforcement learning by latent imagination
  without reconstruction.
\newblock In \emph{2021 IEEE International Conference on Robotics and
  Automation (ICRA)}, pp.\  4209--4215. IEEE, 2021.

\bibitem[Ota et~al.(2020)Ota, Oiki, Jha, Mariyama, and Nikovski]{Ota2020Can}
Ota, K., Oiki, T., Jha, D., Mariyama, T., and Nikovski, D.
\newblock Can increasing input dimensionality improve deep reinforcement
  learning?
\newblock In III, H.~D. and Singh, A. (eds.), \emph{Proceedings of the 37th
  International Conference on Machine Learning}, volume 119 of
  \emph{Proceedings of Machine Learning Research}, pp.\  7424--7433. PMLR,
  13--18 Jul 2020.
\newblock URL \url{https://proceedings.mlr.press/v119/ota20a.html}.

\bibitem[Ota et~al.(2021)Ota, Jha, and Kanezaki]{ota2021training}
Ota, K., Jha, D.~K., and Kanezaki, A.
\newblock Training larger networks for deep reinforcement learning.
\newblock \emph{arXiv e-prints}, pp.\  arXiv--2102, 2021.

\bibitem[Parisotto et~al.(2020)Parisotto, Song, Rae, Pascanu, Gulcehre,
  Jayakumar, Jaderberg, Kaufman, Clark, Noury,
  et~al.]{parisotto2020stabilizing}
Parisotto, E., Song, F., Rae, J., Pascanu, R., Gulcehre, C., Jayakumar, S.,
  Jaderberg, M., Kaufman, R.~L., Clark, A., Noury, S., et~al.
\newblock Stabilizing transformers for reinforcement learning.
\newblock In \emph{International conference on machine learning}, pp.\
  7487--7498. PMLR, 2020.

\bibitem[Paster et~al.(2022)Paster, McIlraith, and Ba]{paster2022you}
Paster, K., McIlraith, S.~A., and Ba, J.
\newblock You can{\textquoteright}t count on luck: Why decision transformers
  fail in stochastic environments.
\newblock In \emph{Decision Awareness in Reinforcement Learning Workshop at
  ICML 2022}, 2022.
\newblock URL \url{https://openreview.net/forum?id=DfCBqPKLsA}.

\bibitem[Raffin et~al.(2021)Raffin, Hill, Gleave, Kanervisto, Ernestus, and
  Dormann]{SB3}
Raffin, A., Hill, A., Gleave, A., Kanervisto, A., Ernestus, M., and Dormann, N.
\newblock Stable-baselines3: Reliable reinforcement learning implementations.
\newblock \emph{Journal of Machine Learning Research}, 22\penalty0
  (268):\penalty0 1--8, 2021.
\newblock URL \url{http://jmlr.org/papers/v22/20-1364.html}.

\bibitem[Schulman et~al.(2015)Schulman, Levine, Abbeel, Jordan, and
  Moritz]{schulman2015trust}
Schulman, J., Levine, S., Abbeel, P., Jordan, M., and Moritz, P.
\newblock Trust region policy optimization.
\newblock In \emph{International conference on machine learning}, pp.\
  1889--1897. PMLR, 2015.

\bibitem[Schulman et~al.(2017)Schulman, Wolski, Dhariwal, Radford, and
  Klimov]{John2017PPO}
Schulman, J., Wolski, F., Dhariwal, P., Radford, A., and Klimov, O.
\newblock Proximal policy optimization algorithms.
\newblock \emph{CoRR}, abs/1707.06347, 2017.
\newblock URL \url{http://arxiv.org/abs/1707.06347}.

\bibitem[Shah \& Kumar(2021)Shah and Kumar]{RRL2021Shah}
Shah, R.~M. and Kumar, V.
\newblock Rrl: Resnet as representation for reinforcement learning.
\newblock In Meila, M. and Zhang, T. (eds.), \emph{Proceedings of the 38th
  International Conference on Machine Learning}, volume 139 of
  \emph{Proceedings of Machine Learning Research}, pp.\  9465--9476. PMLR,
  18--24 Jul 2021.
\newblock URL \url{https://proceedings.mlr.press/v139/shah21a.html}.

\bibitem[Sinha et~al.(2020)Sinha, Bharadhwaj, Srinivas, and
  Garg]{sinha2020d2rl}
Sinha, S., Bharadhwaj, H., Srinivas, A., and Garg, A.
\newblock D2rl: Deep dense architectures in reinforcement learning.
\newblock \emph{arXiv preprint arXiv:2010.09163}, 2020.

\bibitem[Sopov \& Makarov(2022)Sopov and Makarov]{sopov2022transformer}
Sopov, V. and Makarov, I.
\newblock Transformer-based deep reinforcement learning in vizdoom.
\newblock In \emph{International Conference on Analysis of Images, Social
  Networks and Texts}, pp.\  96--110. Springer, 2022.

\bibitem[Sorokin et~al.(2015)Sorokin, Seleznev, Pavlov, Fedorov, and
  Ignateva]{sorokin2015deep}
Sorokin, I., Seleznev, A., Pavlov, M., Fedorov, A., and Ignateva, A.
\newblock Deep attention recurrent q-network.
\newblock \emph{arXiv preprint arXiv:1512.01693}, 2015.

\bibitem[Sutton \& Barto(2018)Sutton and Barto]{sutton2018reinforcement}
Sutton, R.~S. and Barto, A.~G.
\newblock \emph{Reinforcement learning: An introduction}.
\newblock MIT press, 2018.

\bibitem[Takuma~Seno(2021)]{d3rlpy}
Takuma~Seno, M.~I.
\newblock d3rlpy: An offline deep reinforcement library.
\newblock In \emph{NeurIPS 2021 Offline Reinforcement Learning Workshop},
  December 2021.

\bibitem[Tamar et~al.(2016)Tamar, Wu, Thomas, Levine, and
  Abbeel]{tamar2016value}
Tamar, A., Wu, Y., Thomas, G., Levine, S., and Abbeel, P.
\newblock Value iteration networks.
\newblock \emph{Advances in neural information processing systems}, 29, 2016.

\bibitem[Tao et~al.(2022)Tao, Reda, and van~de Panne]{tao2022evaluating}
Tao, T., Reda, D., and van~de Panne, M.
\newblock Evaluating vision transformer methods for deep reinforcement learning
  from pixels.
\newblock \emph{arXiv preprint arXiv:2204.04905}, 2022.

\bibitem[Vaswani et~al.(2017)Vaswani, Shazeer, Parmar, Uszkoreit, Jones, Gomez,
  Kaiser, and Polosukhin]{vaswani2017attention}
Vaswani, A., Shazeer, N., Parmar, N., Uszkoreit, J., Jones, L., Gomez, A.~N.,
  Kaiser, {\L}., and Polosukhin, I.
\newblock Attention is all you need.
\newblock \emph{Advances in neural information processing systems}, 30, 2017.

\bibitem[Wang et~al.(2016)Wang, Schaul, Hessel, Van~Hasselt, Lanctot, and
  De~Freitas]{Wang2016Dueling}
Wang, Z., Schaul, T., Hessel, M., Van~Hasselt, H., Lanctot, M., and De~Freitas,
  N.
\newblock Dueling network architectures for deep reinforcement learning.
\newblock In \emph{Proceedings of the 33rd International Conference on
  International Conference on Machine Learning - Volume 48}, ICML'16, pp.\
  1995–2003. JMLR.org, 2016.

\bibitem[Yu et~al.(2022)Yu, Du, LaLonde, Davila, Funk, Hoogs, and
  Clipp]{Yu2022CascadeTransformers}
Yu, R., Du, D., LaLonde, R., Davila, D., Funk, C., Hoogs, A., and Clipp, B.
\newblock Cascade transformers for end-to-end person search.
\newblock In \emph{Proceedings of the IEEE/CVF Conference on Computer Vision
  and Pattern Recognition (CVPR)}, pp.\  7267--7276, June 2022.

\bibitem[Zheng et~al.(2022)Zheng, Zhang, and Grover]{zheng2022online}
Zheng, Q., Zhang, A., and Grover, A.
\newblock Online decision transformer.
\newblock \emph{arXiv preprint arXiv:2202.05607}, 2022.

\end{thebibliography}
\bibliographystyle{icml2022}

%%%%%%%%%%%%%%%%%%%%%%%%%%%%%%%%%%%%%%%%%%%%%%%%%%%%%%%%%%%%%%%%%%%%%%%%%%%%%%%
%%%%%%%%%%%%%%%%%%%%%%%%%%%%%%%%%%%%%%%%%%%%%%%%%%%%%%%%%%%%%%%%%%%%%%%%%%%%%%%
% APPENDIX
%%%%%%%%%%%%%%%%%%%%%%%%%%%%%%%%%%%%%%%%%%%%%%%%%%%%%%%%%%%%%%%%%%%%%%%%%%%%%%%
%%%%%%%%%%%%%%%%%%%%%%%%%%%%%%%%%%%%%%%%%%%%%%%%%%%%%%%%%%%%%%%%%%%%%%%%%%%%%%%
\newpage
\appendix
\onecolumn

\section{Experiment Details}\label{sec:ExperimentDetails_Appendix}
\subsection{Hyperparameter of PPO\_TIT}
\textbf{Hyperparameters:} For common hyperparameters, we use the exact values proposed by stable baselines3 (SB3), which can be found in: 1) \url{https://stable-baselines3.readthedocs.io/en/master/modules/ppo.html#parameters}; and 2) \url{https://github.com/DLR-RM/rl-baselines3-zoo/blob/master/hyperparams/ppo.yml}. In Table \ref{tab:HyperparameterPPO_Atari_Appendix} and \ref{tab:HyperparameterPPO_Mojoco_Appendix}, we only list the specific hyperparameters for Transformers used in our PPO\_TIT. \emph{Note that PPO\_TIT needs some hypertuning (with common values for the Transformer), but it does not need complex optimization skills for good performance.}

\textbf{Training:} Based on the official results shown in \url{https://github.com/DLR-RM/rl-baselines3-zoo/blob/master/benchmark.md}, we train different networks with PPO for 100,000, 1,000,000 and 10,000,000 timesteps on Classic Control tasks, MuJoCo tasks and Atari tasks, respectively. This is exactly the same as SB3.

\textbf{Evaluation:} Based on the official results shown in \url{https://github.com/DLR-RM/rl-baselines3-zoo/blob/master/benchmark.md}, we find that the official SB3 evaluates the converged agents with 150,000, 150,000, and 600,000 timesteps on Classic Control tasks, MuJoCo tasks and Atari tasks, respectively. However, in our implementation, we evaluate the converged agents with 100 episodes for all tasks. We argue that 100 episodes are enough for creditable evaluation of Gym tasks. As official SB3, we report the mean and the standard deviation of the episode returns of 100 episodes. Our experimental results are based on five independent runs of training and evaluation (seeds are 0, 1, 2, 3, and 4). 

\begin{table*}[h]
\caption{Hyperparameter of PPO\_TIT. The values in this table are used in Atari tasks.}
\label{tab:HyperparameterPPO_Atari_Appendix}
\vskip 0.15in
\begin{center}
\begin{small}
\begin{sc}
\begin{tabular}{ll}
\toprule
Hyperparameter Name & Value \\
\midrule
number of TIT blocks & 2, 4 \\
patch size & 6, 12, 42 \\
embedding dimension & 64, 128 \\
number of attention heads for inner block & 2, 4 \\
number of attention heads for outer block & 2, 4 \\
attention dropout for inner block & 0.1 \\
ffn dropout for inner block & 0.1 \\
attention dropout for outer block & 0.1 \\
ffn dropout for outer block & 0.1 \\
activation function for inner block & gelu \\
activation function for outer block & gelu \\
\bottomrule
\end{tabular}
\end{sc}
\end{small}
\end{center}
\vskip -0.1in
\end{table*}

\begin{table*}[h]
\caption{Hyperparameter of PPO\_TIT. The values in this table are used in MoJoCo and Classic Control tasks.}
\label{tab:HyperparameterPPO_Mojoco_Appendix}
\vskip 0.15in
\begin{center}
\begin{small}
\begin{sc}
\begin{tabular}{ll}
\toprule
Hyperparameter Name & Value \\
\midrule
number of TIT blocks & 2 \\
patch size & 1 \\
embedding dimension & 32, 64 \\
number of attention heads for inner block & 1 \\
number of attention heads for outer block & 1 \\
attention dropout for inner block & 0.0 \\
ffn dropout for inner block & 0.0 \\
attention dropout for outer block & 0.0 \\
ffn dropout for outer block & 0.0 \\
activation function for inner block & gelu \\
activation function for outer block & gelu \\
\bottomrule
\end{tabular}
\end{sc}
\end{small}
\end{center}
\vskip -0.1in
\end{table*}

\subsection{Hyperparameter of CQL\_TIT}\label{sec:CQL_TIT_Hyperparameter}
\textbf{Hyperparameters:} For common hyperparameters, we use the exact values proposed by d3rlpy, which can be found in: 1) \url{https://d3rlpy.readthedocs.io/en/v1.1.1/references/generated/d3rlpy.algos.CQL.html}; and 2) \url{https://github.com/takuseno/d3rlpy/blob/master/reproductions/offline/cql.py#L28}. In Table \ref{tab:HyperparameterCQL_Mojoco_Appendix}, we only list the specific hyperparameters for Transformers used in our CQL\_TIT. \emph{Note that we only tune the values of these hyperparameters about 5 times, and simply set most values based on our empirical experiences, and they work well in offline RL settings.} This demonstrates a key difference between online RL and offline RL.

\begin{table*}[h]
\caption{Hyperparameter of CQL\_TIT.}
\label{tab:HyperparameterCQL_Mojoco_Appendix}
\vskip 0.15in
\begin{center}
\begin{small}
\begin{sc}
\begin{tabular}{ll}
\toprule
Hyperparameter Name & Value \\
\midrule
number of TIT blocks & 2 \\
patch size & observation size \\
embedding dimension & 256 \\
number of attention heads for inner block & 1 \\
number of attention heads for outer block & 4 \\
attention dropout for inner block & 0.0 \\
ffn dropout for inner block & 0.0 \\
attention dropout for outer block & 0.0 \\
ffn dropout for outer block & 0.1 \\
activation function for inner block & gelu \\
activation function for outer block & gelu \\
\bottomrule
\end{tabular}
\end{sc}
\end{small}
\end{center}
\vskip -0.1in
\end{table*}

\textbf{Training:} We use the standard MuJoCo datasets from D4RL \cite{fu2020d4rl} to train the networks with CQL for 500,000 timesteps (see \url{https://github.com/takuseno/d3rlpy/blob/master/reproductions/offline/cql.py#L41}). This is exactly the same as d3rlpy.

\textbf{Evaluation:} As d3rlpy, we evaluate the converged agents every 1 training epoch (which is made up of 1,000 training timesteps, see \url{https://github.com/takuseno/d3rlpy/blob/master/reproductions/offline/cql.py#L42}), so we will get 500,000/1,000=500 evaluation scores. Here, each evaluation score is the average episode\_reward by testing the converged agents with 10 episodes (see \url{https://github.com/takuseno/d3rlpy/blob/master/reproductions/offline/cql.py#L45} and \url{https://github.com/takuseno/d3rlpy/blob/master/d3rlpy/metrics/scorer.py#L405}). We normalize the best evaluation score out of all 500 scores (note that this is the same for the original CQL, see Section \ref{sec:ReferencedResultsD3rlpy} for detials). The normalized score is calculated by $100.0 * (score - random\_score) / (expert\_score - random\_score)$; the random score and expert score are shown in Table \ref{tab:OfflineRandomExpertScore_Mojoco_Appendix}, which can be found in \url{https://github.com/takuseno/d3rlpy-benchmarks/blob/main/d3rlpy_benchmarks/utils.py#L16} and \url{https://github.com/Farama-Foundation/D4RL/blob/master/d4rl/infos.py} (the second link has other MuJoCo tasks' information). We finally report the mean and the standard deviation of the normalized score based on five independent runs of training and evaluation (the running seeds are 0, 1, 2, 3, and 4).

\begin{table*}[h]
\caption{The random score and expert score for MoJoCo tasks.}
\label{tab:OfflineRandomExpertScore_Mojoco_Appendix}
\vskip 0.15in
\begin{center}
\begin{small}
\begin{sc}
\begin{tabular}{lll}
\toprule
Task Name & Random Score & Expert Score \\
\midrule
halfcheetah & -280.178953 & 12135.0 \\
hopper & -20.272305 & 3234.3 \\
walker & 1.629008 & 4592.3 \\
\bottomrule
\end{tabular}
\end{sc}
\end{small}
\end{center}
\vskip -0.1in
\end{table*}

\subsection{Hyperparameter of DT\_TIT}
\textbf{Hyperparameters:} For common hyperparameters, we use the exact values proposed by Decision Transformer (DT), which can be found in: \url{https://github.com/kzl/decision-transformer/blob/master/gym/experiment.py#L283}. For the specific hyperparameters of Transformers used in our DT\_TIT, they are shown in Table \ref{tab:HyperparameterDT_Mojoco_Appendix}. Here, 1) we do not need to set the values of outer Transformer, because the authors implement DT highly based on huggingface \footnote{https://github.com/huggingface/transformers}, and it is not easy to modify the outer Transformer, so we use the default values as DT; 2) the activation function of inner block is set to relu, which is intent to match the default setting of DT: \url{https://github.com/kzl/decision-transformer/blob/master/gym/experiment.py#L295}. \emph{Note that we did not tune the values of these hyperparameters at all, but simply set an empirical values based on our experiences, and they work well in offline SL settings.} This demonstrates a key difference between online RL and offline SL.

\begin{table*}[h]
\caption{Hyperparameter of DT\_TIT.}
\label{tab:HyperparameterDT_Mojoco_Appendix}
\vskip 0.15in
\begin{center}
\begin{small}
\begin{sc}
\begin{tabular}{ll}
\toprule
Hyperparameter Name & Value \\
\midrule
number of TIT blocks & 1 \\
patch size & observation size \\
embedding dimension & 128 \\
number of attention heads for inner block & 1 \\
number of attention heads for outer block & - \\
attention dropout for inner block & 0.0 \\
ffn dropout for inner block & 0.0 \\
attention dropout for outer block & - \\
ffn dropout for outer block & - \\
activation function for inner block & relu \\
activation function for outer block & - \\
\bottomrule
\end{tabular}
\end{sc}
\end{small}
\end{center}
\vskip -0.1in
\end{table*}

\textbf{Training:} We use exactly the same training procedure as Decision Transformer. Specifically, we first download the datasets based on: \url{https://github.com/kzl/decision-transformer/blob/master/gym/data/download_d4rl_datasets.py}, then train the DT and DT\_TIT agents by 10 epochs.

\textbf{Evaluation:} We use exactly the same evaluation procedure as Decision Transformer. Specifically, we evaluate the converged agents based on the target return-to-go (12000 and 6000 for halfcheetah; 3600 and 1800 for hopper; 5000 and 2500 for walker2d) after each training epoch, and report the max return (which has been normalized based on Table \ref{tab:OfflineRandomExpertScore_Mojoco_Appendix}) among 10 epochs. Our experimental results are based on three independent runs of training and evaluation as the original DT (the running seeds are randomly generated as the original DT).

\subsection{The Architecture of DT\_TIT and the Comparison with Other TIT Models}\label{sec:DT_TIT_Architecture}
\begin{figure}[h]
\vskip 0.2in
\begin{center}
\centerline{\includegraphics[width=0.98\columnwidth]{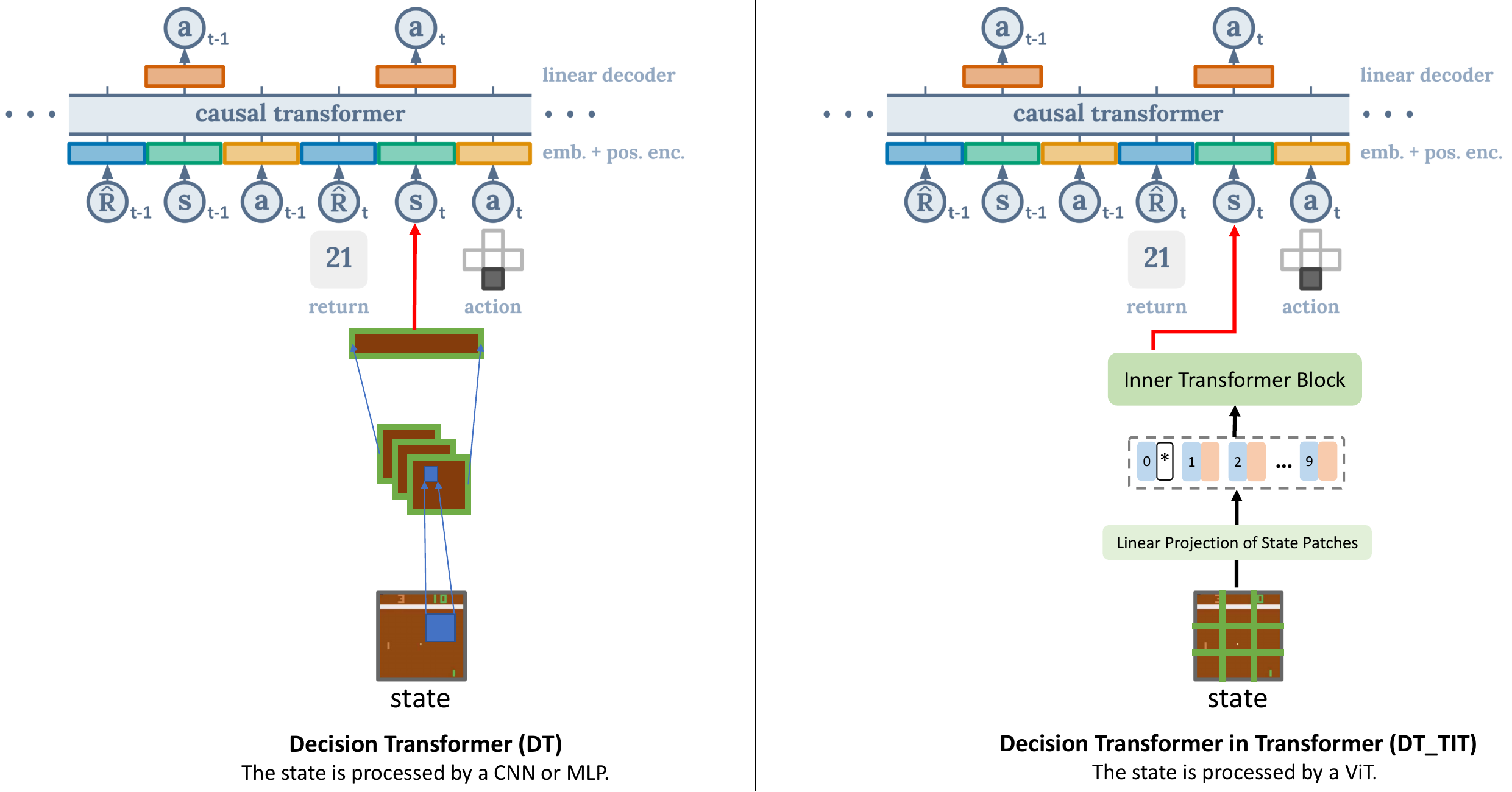}}
\caption{The comparison between the original DT (left) and our DT\_TIT (right).}
\label{fig:DT_TIT_Appendix}
\end{center}
\vskip -0.2in
\end{figure}

The Network Architecture of DT\_TIT is shown on the right of Figure \ref{fig:DT_TIT_Appendix}.

Comparing DT\_TIT with DT, the state is processed by an inner CNN in DT, while the state is processed by an inner ViT in DT\_TIT. In contrast, the sequence of tuples of $\langle observation, action, return \rangle$ is processed by the outer causal transformer.

Comparing DT\_TIT with Vanilla\_TIT shown in Figure \ref{fig:TIT_First}, both have exactly the same inner ViT; however, the outer causal transformer in DT\_TIT is used to process the sequence of tuples of $\langle observation, action, return \rangle$, but the outer causal transformer in Vanilla\_TIT is used to process the sequence of observations.

In summary, there core idea of TIT (no matter Vanilla\_TIT, Enhanced\_TIT nor DT\_TIT) is that we cascade two Transformers in a natural way: the inner one is used to process a single observation at the \emph{observation patch} level, while the outer one is responsible for processing sequential observations (or tuples of $\langle observation, action, return \rangle$ in the paradigm of DT). In this paper, all models based on this general idea are called TIT.

\subsection{The Referenced Results of Different Algorithms on Different Tasks}\label{sec:ReferencedResults_Appendix}
In the experiment part, we have provided the `Reported' results of different algorithms on different tasks, then reproduced these results to ensure that our implementation of algorithms is correct. Here, we give a more clear description about the `Reported' results. 

\subsubsection{Stable Baseline3}
The `Reported' results of stable baseline3 are referenced from this link: \url{https://github.com/DLR-RM/rl-baselines3-zoo/blob/master/benchmark.md}.

\subsubsection{D3rlpy}\label{sec:ReferencedResultsD3rlpy}
The `Reported' results (i.e., $best\_normalized\_return$ and $best\_normalized\_std$) of d3rlpy are referenced from this link: \url{https://github.com/takuseno/d3rlpy-benchmarks/blob/main/d4rl_table.csv}. Every one installed the d3rlpy-benchmarks library \footnote{\url{https://github.com/takuseno/d3rlpy-benchmarks}} can reproduce these results perfectly, by running 
\begin{eqnarray}
score &=& load\_d4rl\_score(algo\_name, env\_name, dataset\_name) \\
best\_normalized\_return &=& np.mean(score.scores.max(axis=1) \\
best\_normalized\_std &=& np.std(score.scores.max(axis=1))
\end{eqnarray}
where the $load\_d4rl\_score$ function is in this link: \url{https://github.com/takuseno/d3rlpy-benchmarks/blob/main/d3rlpy_benchmarks/data_loader.py#L33}.

Please note that the results of d3rlpy-CQL and original-paper-CQL are different, as shown in Table \ref{tab:CQL_Appendix}. Here, original-paper-CQL means that the results are referenced from the original CQL paper \cite{kumar2020conservative}, which are also referenced in the original DT paper \cite{chen2021decision}. In contrast, d3rlpy-CQL means that the results are referenced from the d3rlpy (i.e., the above link). Note again: we use the d3rlpy-CQL's results because 1) d3rlpy-CQL generally achieves better scores than the original-paper-CQL as shown in Table \ref{tab:CQL_Appendix}; 2) our CQL\_TIT is implemented based on d3rlpy-CQL.

\begin{table*}[h]
\caption{The results of d3rlpy-CQL and original-paper-CQL are different.}
\label{tab:CQL_Appendix}
\vskip 0.15in
\begin{center}
\begin{small}
\begin{sc}
\begin{tabular}{ll|cc}
\toprule
Dataset Type & Task Name & original-paper-CQL & d3rlpy-CQL \\
\midrule
\multirow{3}{*}{Medium} & Halfcheetah & 44.4 & 42.6$\pm$0.1 \\
& Hopper & 58.0 & 100.7$\pm$0.3 \\
& Walker2d & 79.2 & 82.8$\pm$1.3 \\
\midrule
\multirow{3}{*}{Medium Replay} & Halfcheetah & 46.2 & 47.1$\pm$0.6 \\
& Hopper & 48.6 & 85.1$\pm$16.2 \\
& Walker2d & 26.7 & 49.6$\pm$5.2 \\
\bottomrule
\end{tabular}
\end{sc}
\end{small}
\end{center}
\vskip -0.1in
\end{table*}

\subsubsection{Decision Transformer}
The `Reported' results of Decision Transformer (DT) are referenced from Table 2 of the original paper \cite{chen2021decision}.

\subsection{Offline Dataset Information}\label{sec:Dataset_Appendix}
We must point out that d3rlpy-CQL and DT use different versions of offline datasets! Specifically, d3rlpy-CQL uses the d4rl-v0 datasets, but DT uses the d4rl-v2 datasets; and the differences between v0 and v2 can be found in this link: \url{https://github.com/Farama-Foundation/d4rl/wiki/Tasks#gym}. In general, the results based on different datasets can NOT be comparable; \textbf{but note that for each separate algorithm, our d3rlpy-CQL\_TIT is comparable with d3rlpy-CQL, and our DT\_TIT is comparable with DT.}

We notice that the DT's authors compare their results with the CQL's results shown in the original paper \cite{kumar2020conservative}; but DT uses d4rl-v2 datasets while the original CQL uses d4rl-v0 datasets. Therefore, their results may not be comparable, which has also been pointed out by other researchers: \url{https://github.com/kzl/decision-transformer/issues/42}.

In summary, the d3rlpy-CQL and the original-paper-CQL use d4rl-v0 datasets, but DT uses d4rl-v2 datasets.

\section{More Analyses of TIT}\label{appendix:MoreAnalyses}
\subsection{Results for Offline RL Setting}
\begin{figure}[h]
\vskip 0.2in
\begin{center}
\centerline{\includegraphics[width=0.46\columnwidth]{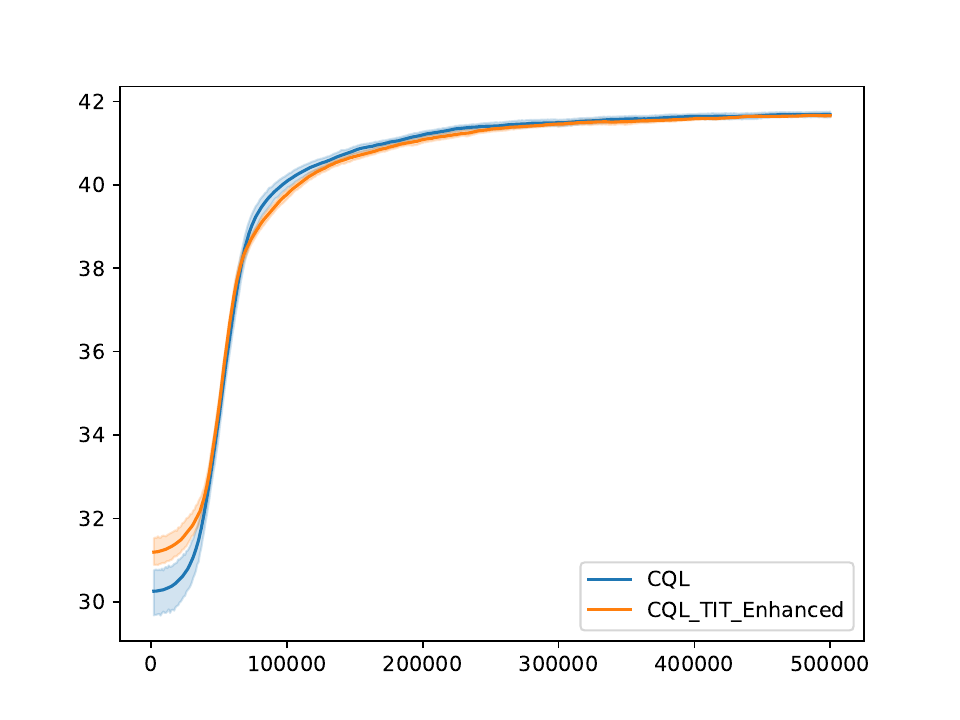}\includegraphics[width=0.46\columnwidth]{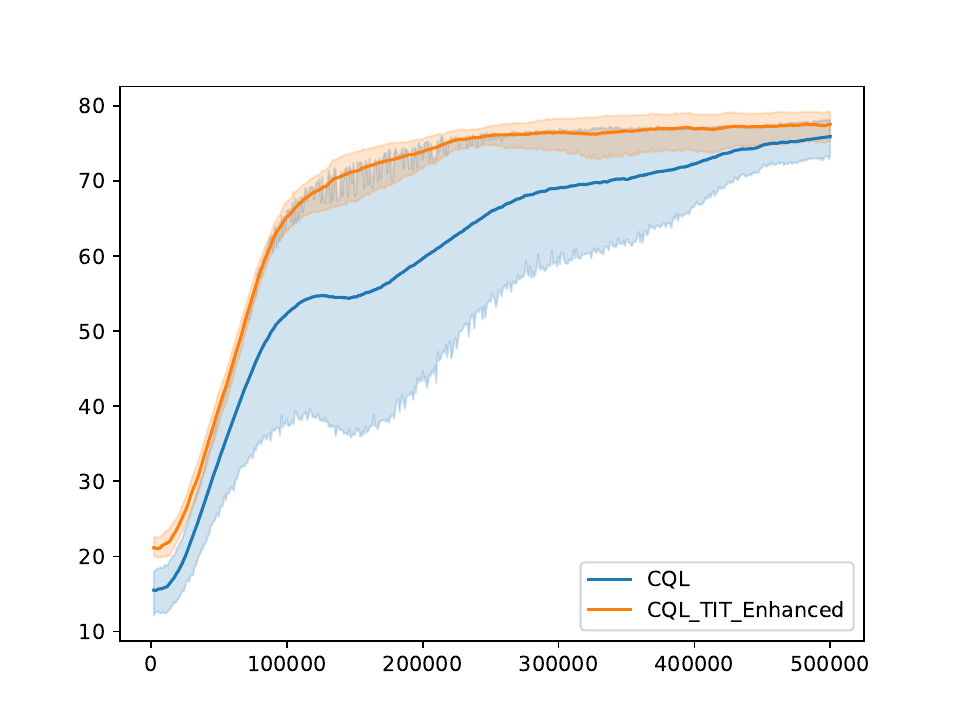}}
\caption{The learning curves on halfcheetah\_medium dataset (left) and walker2d\_medium dataset (right). Although CQL\_TIT and CQL have similar final converged scores, CQL\_TIT has a much faster learning speed than CQL on datasets like walker2d\_medium.}
\label{fig:CQL_TIT_curve}
\end{center}
\vskip -0.2in
\end{figure}

The results in Table \ref{tab:OfflineMain} show that CQL\_TIT works much better than CQL on Medium\_Replay datasets, but it only outperforms CQL by a small margin on Medium datasets. Here, we show the learning curves of different methods in Figure \ref{fig:CQL_TIT_curve}, \textbf{which demonstrates that the CQL\_TIT has a much faster learning speed than CQL on some datasets}, although they have similar final converged scores.

\subsection{Ablation Study}\label{appendix:Ablation}
The ablation results are shown in Table \ref{tab:Ablation_Appendix}. 

The model named `w/o Dense Connection' means that the dense connection design (i.e., the red dotted lines in Figure \ref{fig:TIT_Second}) is removed, and only the output of the last TIT block is used to generate the action. 

The model named `w/o Inner Block' means that the inner blocks are removed, and the input of the outer blocks is the \emph{observation embedding}, which is generated by linearly mapping the whole observation image with an trainable parameter $E^o$, i.e., $y_0 = [o_{t-(K-1)} E^o, o_{t-(K-2)} E^o, ..., o_t E^o]$. 

The model named `w/o Outer Block' means that the outer blocks are removed, and the $K$ observations are stacked in the same way as DQN, then processed by the inner blocks to generate the action.

Compared to Enhanced\_TIT, `w/o Dense Connection' has the maximum performance degradation. \textbf{It implies that the dense connection is important for pure Transformer-based networks}. As far as we know, previous studies have shown that the dense connection is also important for the CNN \cite{Huang2017Densely} and MLP \cite{Ota2020Can,sinha2020d2rl} networks, but TIT is the first work to demonstrate this for Transformer networks. Furthermore, both `w/o Inner Block' and `w/o Outer Block' perform worse than Enhanced\_TIT, which indicates that both inner and outer blocks (and proper arrangement of them) are necessary for good performance.

\begin{table*}[h]
\caption{The results of ablation models in online Atari tasks.}
\label{tab:Ablation_Appendix}
\vskip 0.15in
\begin{center}
\begin{small}
\begin{sc}
\begin{tabular}{l|cccccccc}
\toprule
Task Name & \multicolumn{2}{c}{Breakout} & \multicolumn{2}{c}{MsPacman} & \multicolumn{2}{c}{Pong} & \multicolumn{2}{c}{SpaceInvaders} \\
Obs/Act Space & (1, 84, 84) & 4d & (1, 84, 84) & 9d & (1, 84, 84) & 9d & (1, 84, 84) & 6d \\
\midrule
Episode Return & mean & std & mean & std & mean & std & mean & std \\
\midrule
Vanilla\_TIT & 169 & 91 & 748 & 205 & 9.600 & 6.445 & 752 & 77 \\
Enhanced\_TIT & 321 & 68 & 2246 & 326 & 20.750 & 1.577 & 1645 & 168 \\
\midrule
w/o Dense Connection & 121 & 34 & 1372 & 192 & 18.620 & 3.267 & 938 & 256 \\
w/o Inner Block & 276 & 59 & 1588 & 516 & 19.350 & 2.441 & 1363 & 91 \\
w/o Outer Block & 229 & 83 & 1591 & 396 & 20.180 & 2.034 & 1295 & 233 \\
\bottomrule
\end{tabular}
\end{sc}
\end{small}
\end{center}
\vskip -0.1in
\end{table*}

\subsection{Feature Attention Visualization}\label{appendix:Visualization}
\begin{figure}[t]
\vskip 0.2in
\begin{center}
\centerline{\includegraphics[width=0.98\columnwidth]{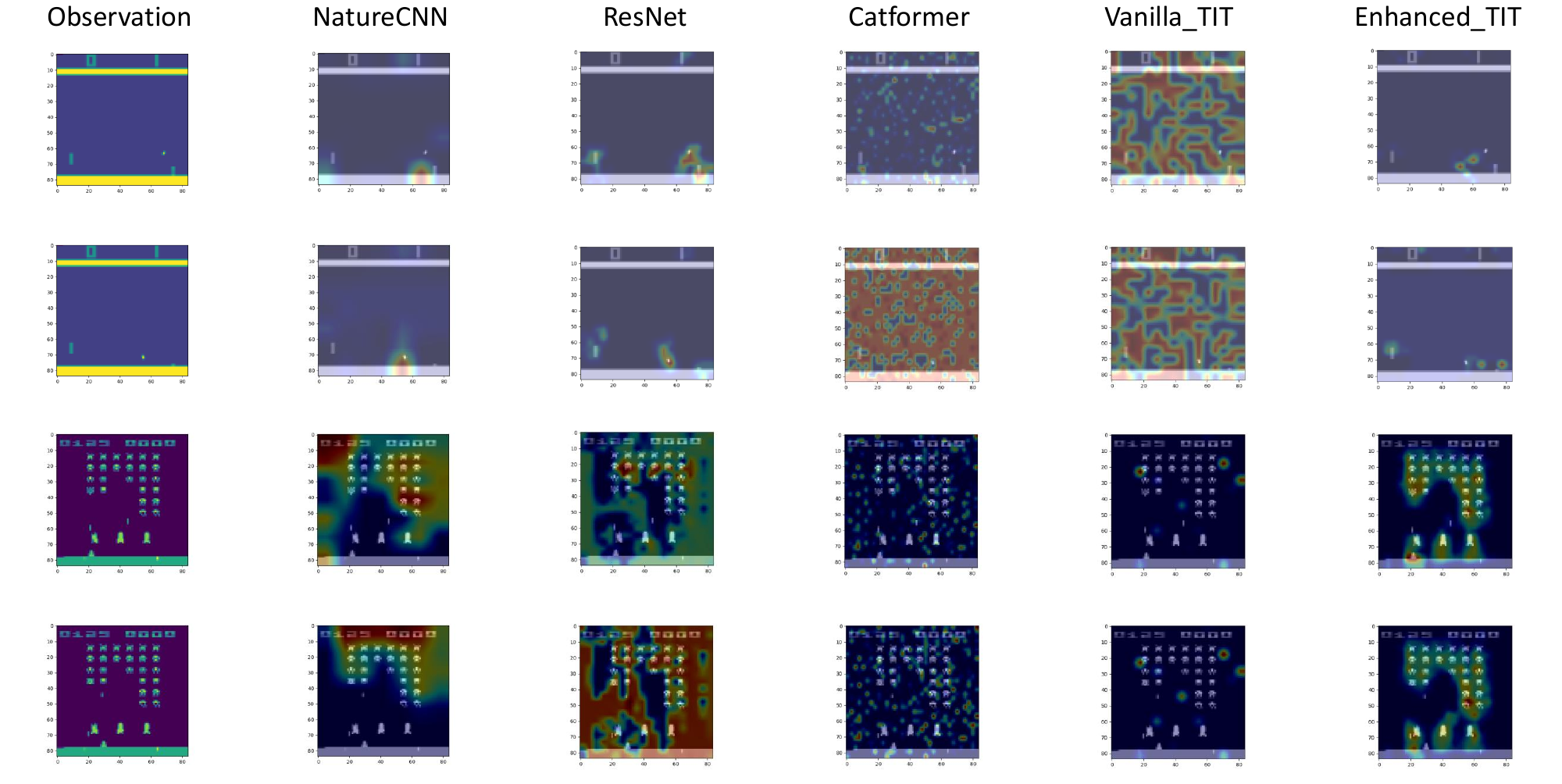}}
\caption{The feature attention visualization of different methods by Grad-CAM. Observations (i.e., the first column) are randomly generated from Pong (i.e., the first two rows) and SpaceInvaders (i.e., the last two rows).}
\label{fig:GradCAM_CompareAppendix}
\end{center}
\vskip -0.2in
\end{figure}

We visualize the feature attention of different methods by Grad-CAM \cite{GradCAM}, which is a typical method for visualization and explainability of deep networks. The results are shown in Figure \ref{fig:GradCAM_CompareAppendix}. 

\textbf{From the spatial perspective, Enhanced\_TIT generates more explainable attention maps than other methods.} As can be seen, the attention of NatureCNN, ResNet and Enhanced\_TIT is highly correlated with the objects in the original observation in most cases. In contrast, Catformer and Vanilla\_TIT have disorganized attention maps, and they sometimes generate unexplainable attention as shown by the first two rows. Furthermore, in the SpaceInvaders task, we found that the attention weight is often positively-correlated with the density of the invaders (i.e., the attention color is darker where there are many invaders). 

\textbf{From the temporal perspective, Enhanced\_TIT generates more stable attention maps than other methods.} For example, when the observation changes slightly (i.e., the second row is slightly changed compared to the first; so does the fourth compared to the third), its attention map does not change significantly. Therefore, Enhanced\_TIT may learn consistent temporal representations for good decision-making, which may explain its stable performance. In contrast, the attention map of NatureCNN and Catformer has changed a lot.

However, please note that the attention map is better used for explanation, but is not necessarily positively-correlated with the performance. For example, Catformer achieves good scores in Pong and SpaceInvaders, but its attention maps are disorganized.

\subsection{Attention Weights Visualization}
\begin{figure}[th!]
\vskip 0.2in
\begin{center}
\centerline{\includegraphics[width=0.98\columnwidth]{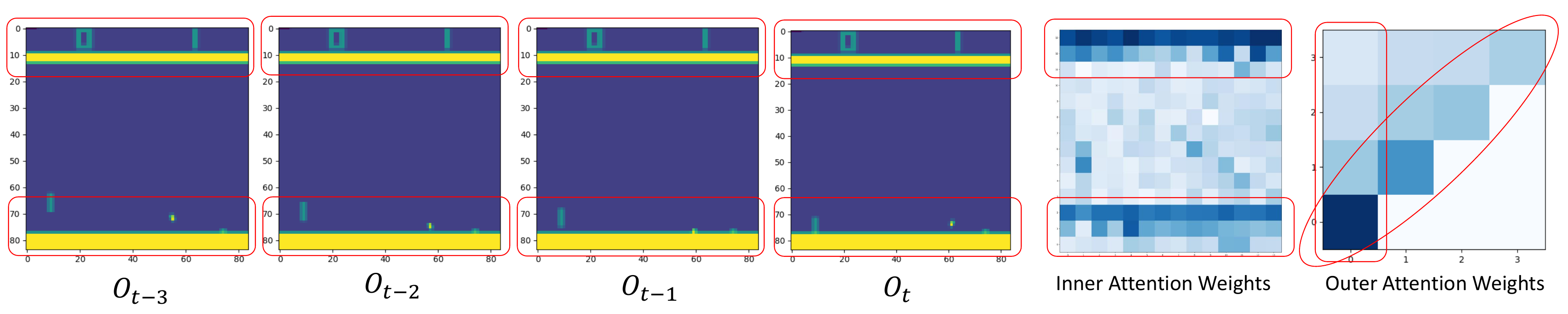}}
\caption{The original attention weights visualization of Enhanced\_TIT. Observations (i.e., the first four images) are randomly generated from Pong. The fifth image is the visualization of attention weights of the Inner Transformer, while the last image is the visualization of attention weights of the Outer Transformer.}
\label{fig:PongAttentionWeights_Appendix}
\end{center}
\vskip -0.2in
\end{figure}

In the above section, we have visualized the \textbf{feature attention} of different methods by \textbf{Grad-CAM}, which cares more about gradients. In this section, we visualize the \textbf{attention weights} directly based on the \textbf{original weight values}. The results are shown in Figure \ref{fig:PongAttentionWeights_Appendix}. 

\textbf{The analyses from spatial perspective.} As shown by the fifth image, which visualizes the attention weights of \emph{the Inner Transformer}. Here, we use the class token as the key in this self-attention operation, because the class token serves as the integrated representation of all observation patches as mentioned in our main paper. As we can see, the top and the bottom parts of the observation image get higher attention weights, since the top part is the game score and the bottom part is the paddles and ball, which are important for the game. In contrast, the middle part of the observation image is mainly the background, which may have little influence on the game, so the attention weight values of the middle part are small. Based on this visualization, we conclude that Enhanced\_TIT (and the Inner Transformer) can indeed capture important spatial information for good decision-making.

\textbf{The analyses from temporal perspective.} As shown by the last image, which visualizes the attention weights of \emph{the Outer Transformer}. Here, we use four observations to form the observation history to match the setting of the original DQN, for fair comparison. As we can see, the current observation $o_t$ usually gets the highest attention weights (i.e., the color on diagonal is darker), so that the most updated information can be used for decision-making. Moreover, the observation (e.g., $o_{t-3}$) far away from the current timestep gets higher attention weights than the middle observations (e.g., $o_{t-1}$ and $o_{t-2}$), therefore the information with the greatest change (in terms of the current observation) can be used for decision-making. Based on this visualization, we conclude that Enhanced\_TIT (and the Outer Transformer) can indeed capture important temporal information for good decision-making.

\section{Discussion and Improvement}\label{appendix:Discussion}
This paper aims at providing a proof of concept that the pure Transformer-based backbones can achieve good results for deep RL, without changing any other settings except for the deep networks. In this section, we give some discussions and point out some directions for further improvements. 

\begin{table}[t]
\caption{The number of information flows for Different TIT backbones.}
\label{tab:InformationFlow}
\vskip 0.15in
\begin{center}
\begin{small}
\begin{sc}
\begin{tabular}{lccc}
\toprule
Flow Type & Vanilla\_TIT & Enhanced\_TIT \\
\midrule
Spatial & LK & LK \\
Temporal & L & L \\
\midrule
S-S & (L-1)K & (L-1)K \\
T-T & L-1 & L \\
S-T & K & LK \\
T-S & 0 & 0 \\
\bottomrule
\end{tabular}
\end{sc}
\end{small}
\end{center}
\vskip -0.1in
\end{table}

\textbf{Information Unit and Information Flow.} It is easy to see that the inner block operates on the observation patches; then the class token integrates the information of all patches, and the outer block operates on this integration. Consequently, there are two types of horizontal information flow: the spatial flow in the inner block and the temporal flow in the outer block. Moreover, there could be four types of vertical information flow: the spatial-to-spatial (S-S) flow, the temporal-to-temporal (T-T) flow, the spatial-to-temporal (S-T) flow, and the temporal-to-spatial (T-S) flow. Table \ref{tab:InformationFlow} shows the number of these flows in different TIT backbones. It can be seen that the Enhanced\_TIT has more information interactions in the vertical direction. 

Besides, comparing Vanilla\_TIT with previous methods like Catformer (i.e., comparing `Transformer + Transformer' with `ResNet + Transformer'), it can be seen that previous methods also operate on observation patches, but they do not have any spatial flows because the CNN in ResNet is local as discussed in ViT \cite{dosovitskiy2020image}.

\textbf{Future Improvement.} On one hand, one can design more advanced network architectures. There are three concrete directions. 1) It could be beneficial if the Transformers directly operate on the \emph{whole observation}. 2) Architectures enabling the T-S information flow can be complementary to the current TIT. 3) The exploration of more advanced Transformers like Swin Transformer \cite{liu2021swin} is also possible. In fact, we tested some advanced architectures, e.g., TNT \cite{han2021transformer} and ViViT \cite{Arnab2021ViViT} from CV community, but did not get consistent improvement. \textbf{We hypothesize that it is inappropriate to train an overly complex architecture only by the signal of RL loss, which motivates our minimal implementation of TIT. Thus, we try to design TIT as simple as possible.}  %  and these shown in Figure \ref{fig:OtherNetworkArchitectures}

% \begin{figure}[t!]
% \vskip 0.2in
% \begin{center}
% \centerline{\includegraphics[width=0.92\columnwidth]{ICML22/figure/TIT_Third.pdf}}
% \caption{Other Transformer in Transformer architectures we tested.}
% \label{fig:OtherNetworkArchitectures}
% \end{center}
% \vskip -0.2in
% \end{figure}

On the other hand, it is exciting to consider dedicated optimization skills to improve the training of pure Transformer-based backbones (although it is not the purpose of this paper), especially for an overly complex backbone. For example, self-supervised methods like next-observation prediction and contrastive representation learning can be very effective to train large Transformer models in RL setting \cite{tao2022evaluating,goulao2022pretraining}.

Besides, Enhanced\_TIT performs slightly worse than the state-of-the-arts in the \textbf{online} MoJoCo tasks. We are currently unable to figure out the exact reasons, but hypothesize that 1) the action space is an important factor for pure Transformer-based RL methods because Classic Control and Atari have discrete action space while MoJoCo has continuous action space; 2) the inner Transformer is actually a ViT, which is good at processing image observation, but MuJoCo has array observation. However, in the \textbf{offline} MuJoCo tasks, our CQL\_TIT is
superior to CQL, and our DT\_TIT is superior to DT. This makes us very confused about these hypotheses since the action space and observation space in the offline MoJoCo setting do not change at all compared to the online MoJoCo setting. This also demonstrates the different working mechanisms between offline RL and online RL algorithms. Therefore, we think the most possible reason is that the offline D4RL dataset is more deterministic than the online MuJoCo environment \cite{paster2022you}, and the deterministic setting is more friendly to the training of large-scale networks like Transformers. We will investigate all of these in the future. % ; 3) the MuJoCo tasks aim to increase the number of independent state that often has a value of zero, which may hinder the learning of our modeling method (recall that we take each entry of state as one patch as mentioned in Section \ref{sec:VanillaInnerTransformer}, so there will be many patch embeddings with zero values)

Finally, we are interested in whether our TIT design is also helpful for Online Decision Transformer \cite{zheng2022online}. Furthermore, since the inner Transformer is a ViT, which is generally superior to CNN, we expect that TIT could demonstrate greater advantages when the observations are more complex or the environment is more partially observable, and we will test this based on DeepMind Lab \cite{beattie2016deepmind} \footnote{\url{https://github.com/deepmind/lab}} which provides a suite of challenging 3D navigation and puzzle-solving tasks for learning agents.

\end{document}